\documentclass[10pt,journal]{IEEEtran} %T-IFS

\usepackage{color}
\usepackage[linesnumbered,boxed,ruled,commentsnumbered]{algorithm2e}
\usepackage{graphicx}
\usepackage{amsmath}
\usepackage[flushleft]{threeparttable}
\usepackage{footnote}
\usepackage{bm}
\usepackage{array}
\usepackage{makecell}
\usepackage{multicol}
\usepackage{multirow}
\usepackage{booktabs}
\usepackage{bigstrut}
\usepackage{mathbbol}
\usepackage{xr}
\usepackage{subfigure}
\usepackage{adjustbox}
\usepackage{mathtools}
\usepackage{soul, color, xcolor}
\usepackage{cite}
\newcommand{\THETA}{\bm{\Theta}}
\newcommand{\PHI}{\bm{\Phi}}

\newcommand{\fig}[1]{Fig.~\ref{#1}}
\newcommand{\tab}[1]{TABLE~\ref{#1}}
\soulregister{\cite}7
\soulregister{\ref}7
\usepackage{algpseudocode}% http://ctan.org/pkg/algorithmicx

\newcolumntype{M}[1]{>{\centering\arraybackslash}m{#1}}
\ifCLASSINFOpdf
\else
\fi

\begin{document}
%
% paper title
% Titles are generally capitalized except for words such as a, an, and, as,
% at, but, by, for, in, nor, of, on, or, the, to and up, which are usually
% not capitalized unless they are the first or last word of the title.
% Linebreaks \\ can be used within to get better formatting as desired.
% Do not put math or special symbols in the title.
% \title{Learning to Extract Meta Texture for Face Anti-Spoofing}
\title{Learning Meta Pattern for Face Anti-Spoofing}

\author{Rizhao~Cai,
    Zhi~Li,
    Renjie~Wan,
    Haoliang~Li$^*$,
    Yongjian~Hu,
    and~Alex~C.~Kot,~\IEEEmembership{Fellow,~IEEE}
\thanks{Rizhao Cai, Renjie Wan, Alex C. Kot are with the ROSE Lab, School of EEE, Nanyang Technological University (email:\{rzcai; rjwan; eackot\}@ntu.edu.sg).}% <-this % stops a space
\thanks{Zhi Li is with the School of Computer Science Engineering, Nanyang Technological University (email: zhi003@e.ntu.edu.sg).}
\thanks{Haoliang Li (*corresponding author) is with the Department of Electrical Engineering, City University of Hong Kong (email:haoliang.li@cityu.edu.hk).}% <-this % stops a space
\thanks{Yongjian Hu is with the South China University of Technology, Guangzhou, China (email: eeyjhu@scut.edu.cn). }

}
% \title{Meta Pattern analysis for Face Anti-Spoofing}
%\title{Texture-Zero: Learning to Learn Meta Pattern for Face Anti-Spoofing}
%
% author names and IEEE memberships
% note positions of commas and nonbreaking spaces ( ~ ) LaTeX will not break
% a structure at a ~ so this keeps an author's name from being broken across
% two lines.
% use \thanks{} to gain access to the first footnote area
% a separate \thanks must be used for each paragraph as LaTeX2e's \thanks
% was not built to handle multiple paragraphs
%

\maketitle
% \thanks{Manuscript received April 19, 2005; revised August 26, 2015.}}

% note the % following the last \textit{i.e. }EEmembership and also \thanks - 
% these prevent an unwanted space from occurring between the last author name
% and the end of the author line. i.e., if you had this:
% 
% \author{....lastname \thanks{...} \thanks{...} }
%                     ^------------^------------^----Do not want these spaces!
%
% a space would be appended to the last name and could cause every name on that
% line to be shifted left slightly. This is one of those "LaTeX things". For
% instance, "\textbf{A} \textbf{B}" will typeset as "A B" not "AB". To get
% "AB" then you have to do: "\textbf{A}\textbf{B}"
% \thanks is no different in this regard, so shield the last } of each \thanks
% that ends a line with a % and do not let a space in before the next \thanks.
% Spaces after \textit{i.e. }EEmembership other than the last one are OK (and needed) as
% you are supposed to have spaces between the names. For what it is worth,
% this is a minor point as most people would not even notice if the said evil
% space somehow managed to creep in.

% The paper headers
\markboth{Journal of \LaTeX\ Class Files,~Vol.~14, No.~8, August~2015}%
{Shell \MakeLowercase{\textit{et al.}}: Bare Demo of IEEEtran.cls for IEEE Journals}
% The only time the second header will appear is for the odd numbered pages
% after the title page when using the twoside option.
% 
% *** Note that you probably will NOT want to include the author's ***
% *** name in the headers of peer review papers.                   ***
% You can use \ifCLASSOPTIONpeerreview for conditional compilation here if
% you desire.

% If you want to put a publisher's ID mark on the page you can do it like
% this:
%\textit{i.e. }EEpubid{0000--0000/00\$00.00~\copyright~2015 IEEE}
% Remember, if you use this you must call \textit{i.e. }EEpubidadjcol in the second
% column for its text to clear the IEEEpubid mark.

% use for special paper notices
%\textit{i.e. }EEspecialpapernotice{(Invited Paper)}

% make the title area

% As a general rule, do not put math, special symbols or citations
% in the abstract or keywords.
\begin{abstract}
%Face anti-spoofing (FAS) is essential to secure face recognition systems and has been extensively studied for several years. Recent FAS methods  handcrafted features as auxiliary information to train neural-network-based discriminators. Despite that the hybrid methods mitigate the generalization problem under domain shift to some extend, the performance relies on the representation capability of auxiliary information. The selection of auxiliary information is still underexplored. This work aims to automatically learn an auxiliary map via a neural network and use it together with original RGB input to develop a generalized discriminator for FAS. We propose a meta-learning framework that consists of a shallow neural network as an auxiliary information extractor and a two-stream neural network as a discriminator. The auxiliary information extractor generates discriminative and generalized auxiliary maps. The discriminator hierarchically fuses the auxiliary map with the original RGB input for inference. Since the training of the framework is a bi-level optimization problem that is non-trivial to solve, we propose an approximation method to simplify the optimization. To verify the effectiveness of proposed method, we conduct extensive experiments under cross-domain settings. The experimental results show that the learned auxiliary information performs better than handcrafted features, and our proposed method achieves state-of-the-art performance on different domain generalization evaluation benchmarks.
Face Anti-Spoofing (FAS) is essential to secure face recognition systems and has been extensively studied in recent years.  Although deep neural networks (DNNs) for the FAS task have achieved promising results in intra-dataset experiments with similar distributions of training and testing data, the DNNs' generalization ability is limited under the cross-domain scenarios with different distributions of training and testing data. To improve the generalization ability, recent hybrid methods have been explored to extract task-aware handcrafted features (e.g., Local Binary Pattern) as discriminative information for the input of DNNs. However, the handcrafted feature extraction relies on experts' domain knowledge, and how to choose appropriate handcrafted features is underexplored. To this end, we propose a learnable network to extract Meta Pattern (MP) in our learning-to-learn framework. By replacing handcrafted features with the MP, the discriminative information from MP is capable of learning a more generalized model.
Moreover, we devise a two-stream network to hierarchically fuse the input RGB image and the extracted MP by using our proposed Hierarchical Fusion Module (HFM). We conduct comprehensive experiments and show that our MP outperforms the compared handcrafted features. Also, our proposed method with HFM and the MP can achieve state-of-the-art performance on two different domain generalization evaluation benchmarks.

\end{abstract}

% Note that keywords are not normally used for peerreview papers.
%\begin{IEEEkeywords}
%Face anti-spoofing
%\end{IEEEkeywords}

%}

% make the title area
\maketitle

% To allow for easy dual compilation without having to reenter the
% abstract/keywords data, the \textit{i.e. }EEtitleabstractindextext text will
% not be used in maketitle, but will appear (i.e., to be "transported")
% here as \textit{i.e. }EEdisplaynontitleabstractindextext when the compsoc 
% or transmag modes are not selected <OR> if conference mode is selected 
% - because all conference papers position the abstract like regular
% papers do.
% \textit{i.e. }EEdisplaynontitleabstractindextext
% \textit{i.e. }EEdisplaynontitleabstractindextext has no effect when using
% compsoc or transmag under a non-conference mode.

% For peer review papers, you can put extra information on the cover
% page as needed:
\ifCLASSOPTIONpeerreview
\begin{center} \bfseries EDICS Category: 3-BBND \end{center}
\fi
%
% For peerreview papers, this IEEEtran command inserts a page break and
% creates the second title. It will be ignored for other modes.
% \textit{i.e. }EEpeerreviewmaketitle

\section{Introduction}\label{Sec-1}
% What and why Face Anti-Spoofing
\IEEEPARstart{F}{ACE} recognition techniques have been used in various identity authentication scenarios and have become increasingly prevalent in recent years. Despite the ease of use, face recognition systems are vulnerable to Presentation Attacks (PAs), \textit{a.k.a.} Spoofing Attacks, where an attacker may create face forgeries such as printed photos, digital displays, masks, and further launch spoofing attacks by presenting the forgeries to the camera sensor of face recognition systems. To secure face recognition systems, both the industry and academia have been paying increasing attention to the problem of Face Presentation Attack Detection (Face PAD), \textit{a.k.a.} Face Anti-Spoofing (FAS), which aims to discriminate spoofing attacks from bona fide attempts of genuine users. % CHECKED

\begin{figure}[t]
    \includegraphics[width=\linewidth]{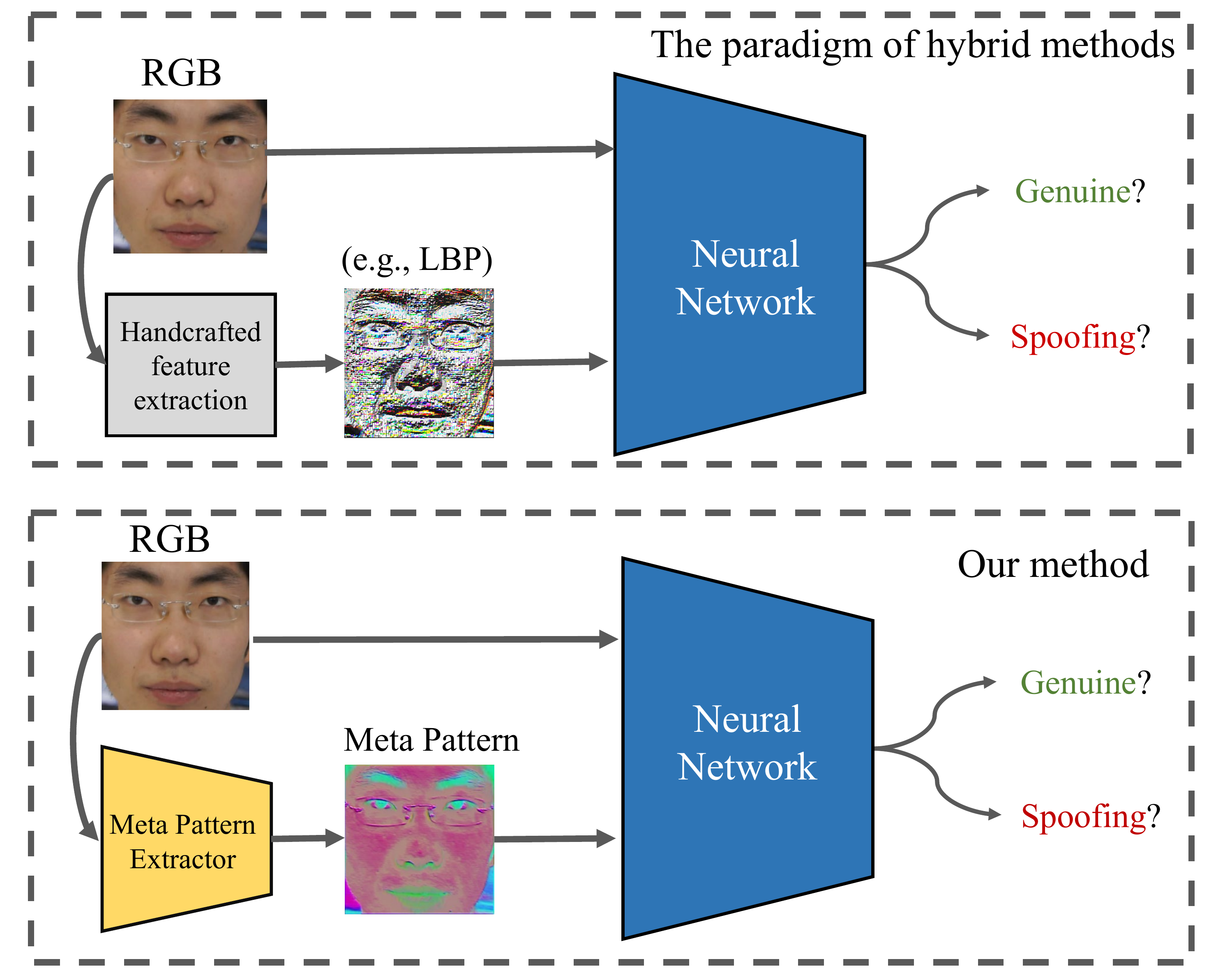}
    \caption{The upper diagram shows the paradigm of typical hybrid methods, which manually extract handcrafted features and combine the handcrafted features (\textit{e.g.,} LBP \cite{hashemifard2021compact}) with neural networks. The bottom diagram illustrates our proposed method of Meta Pattern. We train a Meta Pattern Extractor to extract Meta Pattern to replace handcrafted features.} \label{fig:framework}
\end{figure}
% Inherent disparities
The research community of FAS believes that there are intrinsic disparities between face images captured from bona fide attempts (abbreviated as genuine faces) and spoofing attacks (abbreviated as spoofing faces) \cite{FAS-ColorTexture-TIFS-2016}. For example, digital displays are made of glass and have high reflection coefficients, some texture patterns like reflection can be observed in spoofing faces of replay attacks. Also, for printed photos attacks, the spoofing faces tend to present lower image quality due to the low Dots Per Inch (DPI) and color degradation \cite{FAS-IDA-TIFS-2015}.

Based on observations and analysis of domain knowledge, earlier researchers \cite{FAS-ColorTexture-TIFS-2016, moire-analysis-TIFS-2015,IQA-ICPR-2014,DoG-ECCV-2010, SURF-SPL-2017, LBP-TOP-EJIVP-2014, LBP-FAS-BIOSIG-2012} design various kinds of handcrafted features to describe the disparities between genuine and spoofing faces and leverage the handcrafted features to detect spoofing attacks. The different handcrafted features are usually devised based on different specific physical meanings. For example, handcrafted image descriptors such as Local Binary Pattern (LBP) \cite{LBP-FAS-BIOSIG-2012} and Speeded Up Robust Features (SURF) \cite{SURF-SPL-2017} are used to describe the texture discrepancy between genuine and spoofing faces. Besides, handcrafted features based on image quality have been proposed to detect spoofing attacks by analyzing the image quality (\textit{e.g.}, blurring) \cite{FAS-IDA-TIFS-2015}. Other than analyzing a single frame, handcrafted features from sequential frames have also been proposed to analyze the motion difference between spoofing faces from the genuine ones in the temporal domain \cite{Motion-IJCB-2011, MotionLBP-ICB-2013, LBP-TOP-EJIVP-2014}. 
Despite elegant interpretability, the handcrafted features heavily rely on experts' domain knowledge. These features are devised on the basis of some specific considerations, and few of them can deal with diverse types of attacks.
% Deep learning: end-to-end, without any restriction.
% Only using label 1 can be overffitig

Recently, deep neural networks have also been used to learn discriminative feature representations in a data-driven manner for the FAS problem \cite{FAS-3DCNN-TIFS-2018, CAI-2020-DRL, CDCN-CVPR-2020, FAS-LSTMCNN-ICASSP-2018, LIZHI, Ternary-TIFS-2018, CBL}. The deep learning based methods have outperformed the traditional methods, achieving desired performance in intra-domain experiments.  However, training deep neural networks with only RGB images as input and simple binary labels as supervision easily make the models overfit to the properties of the source domain training data. This results in poor generalization performance in cross-domain experiments where the domain shift between training and testing data exists, including but not limited to the variations of environmental illumination, camera specifications, and materials of spoofing attack mediums \cite{FAS-UnsupervisedDA-TIFS-2018, FAS-3DCNN-TIFS-2018}. 

Among various techniques that aim to mitigate the domain shift problem and improve generalization performance, one promising research direction is to combine task-aware handcrafted features and deep neural networks, and such methods are summarized as hybrid methods \cite{FAS-Survey-arXiv-2021}. As illustrated in the top diagram of Fig.~\ref{fig:framework}, some hybrid methods provide discriminative information by extracting handcrafted features from RGB images (\textit{e.g.}, using LBP \cite{hashemifard2021compact}) or transforming the images in the RGB space to other spaces (\textit{e.g.}, the HSV space  \cite{FAS-Auxiliary-CVPR-2018, DTL, LIZHI}) as the input to the neural network models. Although the hybrid strategy can improve the generalization performance of the models to some extent, the extracted handcrafted features may not be representative and generalized enough under complex situations due to the diverse factors of domain shift problem (e.g., different cameras, lightings, attack mediums). As such, how to extract and utilize the features to improve the models' generalization ability poses a unique challenge for the FAS problem.

In this work, we address the above challenge in a learning-to-learn framework. We devise a learnable neural network $\PHI$ to extract the Meta Pattern (MP). The MP replaces handcrafted features to provide discriminative or auxiliary information to the target discriminator network $\THETA$ to discriminate spoofing attacks (see the bottom of Fig.~\ref{fig:framework}).  We expect the learnable network $\PHI$ can extract representative Meta Pattern, which benefits the generalization ability of the discriminator network $\THETA$. To be specific, a bi-level optimization problem is induced, where the optimizations of $\PHI$ and $\THETA$ are in the inner and outer optimization levels respectively. Since solving a bi-level optimization problem is non-trivial, we simplify the optimization by a neat and effective approximation method, which can be easily solved via asynchronous backward propagation (see Section~\ref{Sec-3}) and no surrogate model is needed. Moreover, we devise a two-stream Hierarchical Fusion Network (HFN) to fuse the original information from the RGB images and discriminative information from the MP by using our proposed Hierarchical Fusion Module (HFM). The illustrations of our HFN and HFM are shown in \fig{fig:HFN}.

% \begin{figure}
%    \includegraphics[width=1.0\linewidth]{texture_maps.png}
%    \caption{Different "texture" maps. The top and bottom rows show the different texture maps of the genuine face and fake face respectively. (a) is genuine face, and (b), (c), (d), (e), (d, pseudo) (f, pseudo) are the depth, albedo, reflectance, MSR, and the proposed Meta Pattern of (a). (g), (h), (i), (j), (k, pseudo), and (l, pseudo) are the counterparts of (b). We can see that difference between genuine and fake faces of different maps can also be describe in texture, and such that these "texture" maps could be regarded as "generalized" texture maps.}\label{fig:tex}
%\end{figure}   

We summarize our contributions as follows:
\begin{itemize}
  \item We push the hybrid method one step further to be the end-to-end data-driven method by learning to extract the MP from data, instead of extracting handcrafted features manually. 
  \item We devise a Hierarchical Fusion Network to fuse information from multiple feature hierarchies with our Hierarchical Fusion Module.
  \item We conduct extensive experiments to verify the effectiveness of the proposed method. The experimental results show that our learned MP can generally achieve better performance over the compared handcrafted features, and our proposed method can achieve state-of-the-art performance in the cross-domain generalization benchmarks.
\end{itemize}

The remainder of the paper is as follows. Section \ref{Sec-2} discusses the literature related to our work to provide background information. Section \ref{Sec-3} illustrates our method of learning the MP and presents the details of the HFN. Section \ref{Sec-4} presents the experiments about the preliminary study and ablation study. In Section \ref{Sec-5}, we conclude this paper.

\section{Related works}\label{Sec-2}
% The research community of FAS believes that there are intrinsic disparities between genuine faces and spoofing faces captured by visible light (VIS) cameras. 
Multifarious FAS methods based on RGB images have been proposed in the past decade and have become the mainstream of the FAS research. In this section, we firstly review the development of FAS methods, from traditional methods based on handcrafted features to recent methods based on deep learning. Besides, we also review recent progress of domain generalization methods for FAS, which are most relevant to our work.

\subsection{Traditional methods for face anti-spoofing}
Since spoofing faces of printed photos attacks and replay attacks have undergone multiple capture processes, some texture differences such as blurring, moire patterns, and printed noise can be observed in the spoofing faces because of the image quality distortion during the recapturing \cite{MicroTexture-IJCB-2011, FAS-IDA-TIFS-2015}. Traditional methods use image descriptors such as Local Binary Pattern (LBP), Scale-invariant feature transform (SIFT), Speeded-up Robust Features (SURF), Histogram of Oriented Gradients (HOG), and Difference of Gaussians (DoG) to extract handcrafted features about texture information \cite{LBP-FAS-BIOSIG-2012, SURF-SPL-2017, HOG, DoG-ECCV-2010, deep-forest-lbp}. Considering the color distortion of spoofing faces, Boulkenafet \textit{et al.} \cite{FAS-ColorTexture-TIFS-2016} utilize the color information and propose to extract color texture features from the illuminance and chrominance components (respective channels of images in YCrCb or HSV color spaces). Specific texture analyses for moire patterns have also been studied \cite{moire-icb, moire-analysis-TIFS-2015}. Wen \textit{et al.} \cite{FAS-IDA-TIFS-2015} argue that texture features contain information about personal identity, which is redundant for anti-spoofing and could lead to poor generalization performance. Therefore, some works propose handcrafted features based on image quality and distortion analysis for the FAS task \cite{FAS-IDA-TIFS-2015, IQA-ICPR-2014, LI-QUALITY}. Besides, some methods \cite{Motion-IJCB-2011, MotionLBP-ICB-2013, LBP-TOP-EJIVP-2014} extract dynamic texture features from multiple frames to analyze the motion information in the temporal domain other than in the spatial domain.

\subsection{Deep learning methods for face anti-spoofing}
Deep neural networks show strong feature learning capacity and have been widely used in recent FAS methods. Yang \textit{et al.} \cite{FAS-CNN-ComputerScience-2014} are the first to propose a method that using a VGG-Net \cite{vggnet} as a feature extractor for the FAS task. They firstly extract deep features from the fully connected layer of the VGG-Net, then train a Support Vector Machine (SVM) classifier with the deep features to discriminate genuine and spoofing faces. Motivated by the great success of deep learning techniques,  more FAS methods based on deep learning have been proposed. Cai \textit{et al.} \cite{CAI-2020-DRL} tackle the FAS problem under a reinforcement learning framework. From the perspective of anomaly detection and metric learning, Li \textit{et al.} devise a loss function for optimization of the discrimination network \cite{LIZHI}. Some methods are proposed to use pixel (patch)-wise labels for supervision of network training \cite{FAS-Auxiliary-CVPR-2018, Ternary-TIFS-2018, deeppixel--ICB-2019, yu2021revisiting}. Yu \textit{et al.} design spoofing-aware convolution structures for fine-grained feature learning \cite{NASFAS-TPAMI-2020, CDCN-CVPR-2020}.

However, learning deep models with only RGB images as input will easily make models overfit the training data thus can not generalize well to the testing data if there are domain shifts between training and testing data domains. %Although using other modalities other than RGB images can improve the performance \cite{george2019biometric}, the acquiring other modalities needs special sensors, which could be expensive and not always available. Meanwhile, handcrafted features, such as LBP, optical flow, are shown provide discriminative information for face anti-spoofing \cite{FAS-Survey-arXiv-2021}. 
Recently, some hybrid methods that combine handcrafted features and neural networks have been proposed to improve deep models' generalization performance. For example, Chen \textit{et al.} \cite{FAS-MSR-TIFS-2019} transform images from the RGB space to the illumination-invariant multi-scale retinex (MSR) space and train an attention-based two-stream network with the RGB and MSR. Yu \textit{et al.} construct the spatio-temporal remote photoplethysmography (rPPG) map to represent the signal of heartbeats, and train a vision transformer to detect 3D mask attacks. Li \textit{et al.} \cite{TIFS-2019-MotionBlur} propose to analyze motion blur from replay attacks by fusing features extracted by 1D CNN and Local Similarity Pattern (LSP). Pinto \textit{et al.} \cite{Pinto} consider that the attack mediums and human skins are different. As such, Pinto \textit{et al.} \cite{Pinto} utilize Shape-from-Shading (SfS) algorithm to extract albedo, depth, and reflectance maps as the input of the proposed SfSNet to analyze the material difference between genuine faces and spoofing faces. Some other hybrid methods that take advantage of handcrafted features (\textit{e.g.,} LBP, HOG) and neural networks have also been studied \cite{CNN-LBPTOP-2017, rehman2019perturbing}.

Although these hybrid methods improve the generalization performance of deep neural networks, the handcrafted feature used in each hybrid method is based on special considerations. Due to the various factors of domain shifts, it is hard to consider all possible spoofing information by a type of handcrafted feature. This limitation may constrain the hybrid methods as it would be hard to choose the desired handcrafted features when given different source data. Thus, we push the hybrid methods one step further to end-to-end data-driven methods by learning to extract the Meta Pattern in this work.

\subsection{Domain generalization methods for face anti-spoofing}
The data collection conditions of training and testing data could be different, including but not limited to the variations of camera specifications, environment illuminations, and presentation mediums. Such variations of capture condition result in the shift between training and testing data domain and deter the models' reliability from being deployed in practical scenarios \cite{FAS-UnsupervisedDA-TIFS-2018}. To tackle this problem, domain adaptation methods \cite{FAS-UnsupervisedDA-TIFS-2018, LI-DISTILLATION} utilize some unlabeled target domain data to adapt a model trained with the source domain to improve the performance in the target domain. However, domain adaptation methods require using some target domain data, which is expensive to collect in real-world applications. By contrast, since domain generalization aims to learn a model that can be more generalized to the unseen data domains without using the target domain data \cite{FAS-3DCNN-TIFS-2018}, the FAS methods using domain generalization techniques have been extensively studied in recent years \cite{RFMetaFAS-AAAI-2020, NASFAS-TPAMI-2020, MetaTeacher-TPAMI-2021, FAS-3DCNN-TIFS-2018, MADDG-CVPR-2019}. Meta-learning as an effective way to tackle the general domain generalization problem has been introduced to the face anti-spoofing task. Shao \textit{et al.} use meta-learning to regularize the gradient calculated from the pixel map supervision \cite{RFMetaFAS-AAAI-2020}. Yu \textit{et al. } \cite{NASFAS-TPAMI-2020} use meta-learning and Neural Architecture Search to search the network architecture that can be generalized to unseen data domains. Yu \textit{et al.} \cite{MetaTeacher-TPAMI-2021} propose Meta-Teacher to learn a teacher network that can provide data-specific labels for supervision in a teacher-student framework. Different from existing methods, in this paper, we propose a novel method to learn a domain-generalized model by learning to extract the MP from a Meta Pattern Extractor to provide discriminative information in our meta-learning paradigm. % \hl{(Add more comparisons?)}
% No idea

% Meta-learning, categories
% Meta-learning methods used in face anti-spoofing
\begin{figure*}
    \centering
    \includegraphics[width=0.98\linewidth]{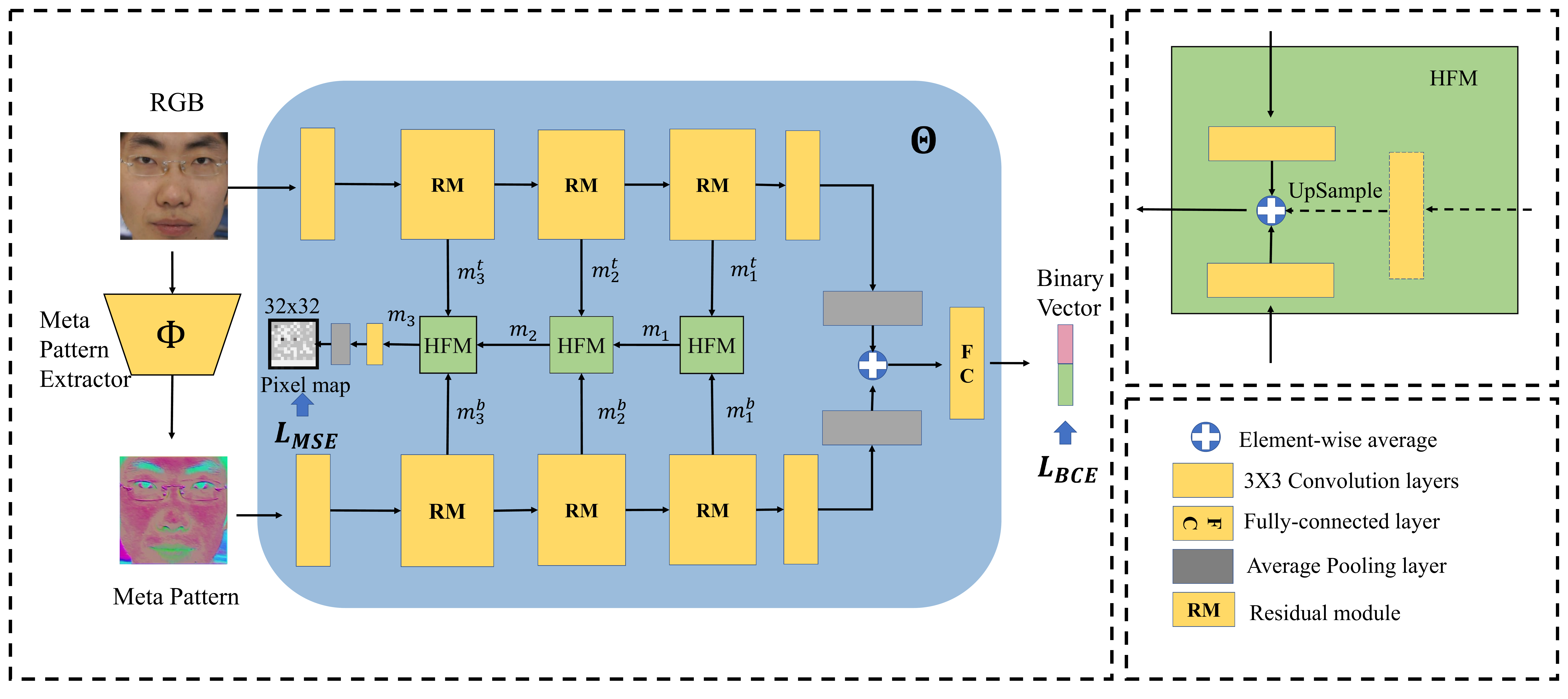}
    \caption{The illustration of the two-stream Hierarchical Fusion Network. ``RM" represents the residual module from ResNet \cite{ResNet}, ``HFM" represents our Hierarchical Fusion Module. The input RGB image is forwarded to the Pattern Extractor to generate the Meta Pattern. Then, the RGB image and Meta Pattern are forwarded to the top stream and the bottom stream respectively. Features from different hierarchies of the two streams are fused via the HFMs. The Binary Classification Entropy Loss ($\mathcal{L}_{BCE}$) is applied to the output binary vector from the fully connected layer, and the pixel-wise Mean Square Error loss ($\mathcal{L}_{MSE}$) is applied to the output pixel map from the final HFM. The dashed line in the diagram of HFM means there is no input feature from the previous hierarchy 0 ($m_0$).}
    \label{fig:HFN}
\end{figure*}
\section{Methodology}\label{Sec-3}
% Overview
%As introduced in previous sections, our method aims to learn a Meta Pattern Extractor $\PHI$ from the training data and use $\PHI$ to extract the Meta Pattern (MP) as the discriminative information to train a generalized discriminator network $\THETA$. 
In this section, we formulate the learning of $\PHI$ and $\THETA$ as a bi-level optimization problem and describe how we solve the optimization problem by using a neat but effective approximation method. After that, we introduce the instantiation details about the Meta Pattern Extractor $\PHI$ and the discriminator network $\THETA$ (Hierarchical Fusion Network), respectively.

%\textcolor{red}{HL: it would be better to remove the sentences before in this section.}

\subsection{Learning to extract Meta Pattern}
\subsubsection{Problem formulation}
The FAS problem can be formulated as a binary classification problem: genuine or spoofing. In general, a neural network $\THETA$ can be can trained for the genuine/spoofing classification and the optimization of $\THETA$ can be solved by Empirical Risk Minimization (ERM) over the training data, which can be expressed as:
    \begin{equation}\label{eq-1}
  		%\begin{aligned}
	  		\mathop{\arg\min}_{\THETA} \ \ \ \mathbb{E}_{x,y\sim D^{S}} \mathcal{L}[ \THETA(x), y],
  		%\end{aligned}
  	\end{equation}
where $x$ is the input RGB image, $y$ is the target label,  $D^{S}$ denotes the source domain data, $\THETA(x)$ denotes the network output given the input $x$, and $\mathcal{L}$ is the loss function (\textit{e.g.}, Cross Entropy loss). For simplicity, hereafter, we rewrite the loss function calculation $\mathcal{L}[\THETA(x), y ]$ used in Eq.~\ref{eq-1} as: 
    \begin{equation}\label{eq-2}
  		%\begin{aligned}
	  	     \mathcal{L}(x, y| \THETA) \ \coloneqq \ \mathcal{L}[\THETA(x), y ].
  		%\end{aligned}
  	\end{equation}

Eq.\ref{eq-1} can be solved by standard Stochastic Gradient Descent (SGD, without the momentum). At $t$-th iteration, given a batch of data $\{x^t, y^t\}$, the update of the $\THETA$ can be expressed as
    \begin{equation}\label{eq-3}
  		%\begin{aligned}
	  	    \THETA^{t} = \THETA^{t-1} - \alpha \nabla_{\THETA}  \mathcal{L}(x^t, y^t|\THETA^{t-1}),
  		%\end{aligned}
  	\end{equation}
where $\alpha$ is the learning rate. To improve the generalization ability, existing hybrid methods extract task-aware handcrafted features as the input to neural networks. If $f$ denotes method (non-learnable) used to extract handcrafted features $f(x)$ in hybrid methods, the optimization objective can be expressed as 
\begin{equation}\label{eq-4-0}
  		%\begin{aligned}
	  		\mathop{\arg\min}_{\THETA} \ \ \ \mathbb{E}_{x,y\sim D^S} \mathcal{L}(x, y| \THETA, f),
  		%\end{aligned}
  	\end{equation}
and the gradient for the update in Eq.~\ref{eq-3} can be rewritten as
\begin{equation}\label{eq-4}
  		%\begin{aligned}
	  	    \THETA^{t} = \THETA^{t-1} - \alpha \nabla_{\THETA}  \mathcal{L}(x^t, y^t|\THETA^{t-1}, f).
  		%\end{aligned}
  	\end{equation}
However, manually devising $f$ could be tricky due to the complexity of the source training data. Therefore, we propose to utilize the learning-to-learn paradigm to replace $f$ with a learnable Meta Pattern Extractor $\PHI$, which is implemented by a convolutional neural network, to extract the Meta Pattern (MP). In this paradigm, we expect that the extracted MP can help $\THETA$ be more generalized. Therefore, the optimization objective of $\PHI$ is to improve the generalization ability of $\THETA$ to unseen data domains. Based on Eq.~\ref{eq-4}, a bi-level optimization problem is induced:
    \begin{equation}\label{eq-5}
  		\begin{array}{cc}
  		     & \THETA^* = \mathop{\arg\min}_{\THETA} \mathbb{E}_{x,y\sim D^S} \mathcal{L}(x, y|\THETA, \PHI^*),\\  \\
  		     & s.t. \ \PHI^* = \mathop{\arg\min}_{\PHI} \mathbb{E}_{x,y\sim D^T} \mathcal{L}(x, y|\THETA, \PHI),
  		\end{array}
  	\end{equation}
where $D^T$ represents unseen domain data. In Eq.~\ref{eq-5}, $\PHI$ is in the inner level instead of the upper level because the $\PHI$ is trained to provide MP as the discriminative and auxiliary information, while $\THETA$ is the target model to be optimized for the face anti-spoofing. Therefore, $\THETA$ is the optimization target in the upper level while $\PHI$ is the optimization target in the inner level.
\begin{figure}[t]
\centering
\includegraphics[width=\linewidth]{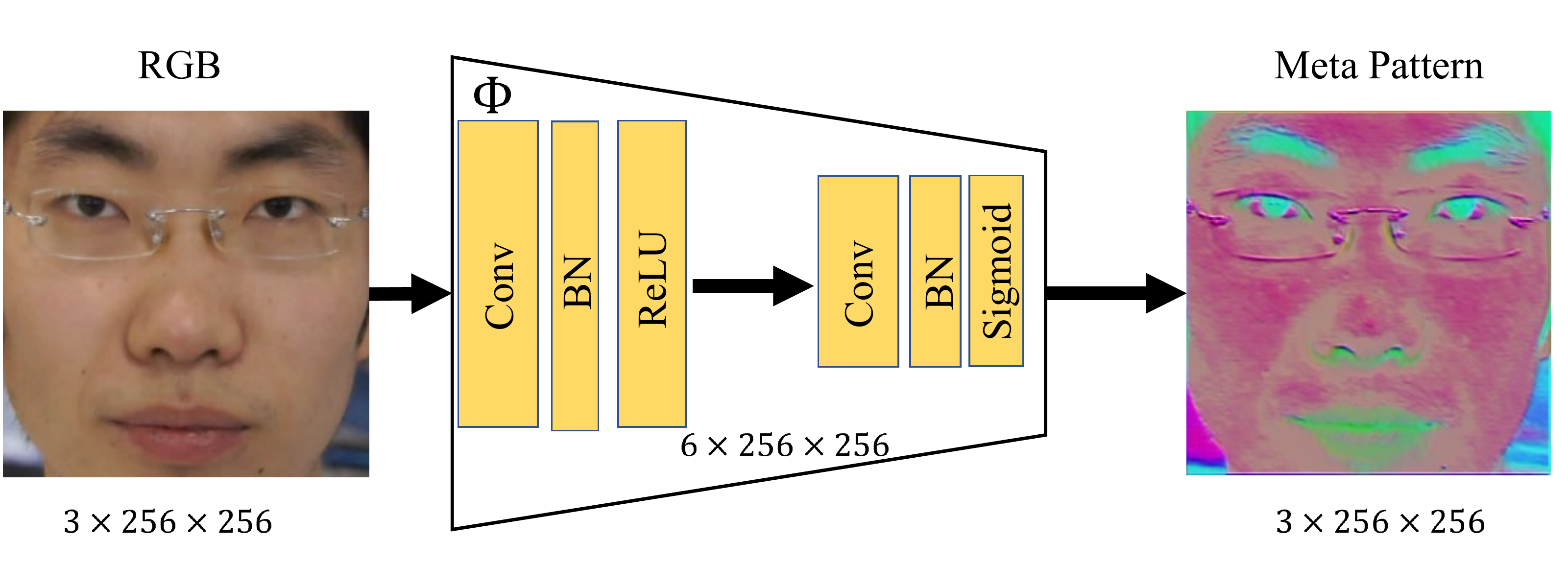}
\caption{The structure of the two-layer Meta Pattern Extractor. ``Conv'' means a vanilla 3$\times$3 convolutional operation, with both padding size and stride as 1. "BN" represents the Batch Normalization layer. "ReLU" and "Sigmoid" represent Rectified Linear Unit and Sigmoid activation functions respectively. }
\label{fig:phi}
\end{figure}
\subsubsection{Solving the bi-level optimization by approximation}
Eq.~\ref{eq-5} illustrates the induced bi-level optimization problem, but solving a bi-level optimization based on gradients is nontrivial as the gradient calculation could be inexplicit and complicated. To solve the original bi-level optimization problem, we propose a neat and effective solution by approximating $\PHI^*$ with a local minima $\hat{\PHI}$, which can be done with several steps of backward propagation.

%The proposed approximation is inspired by recent study of few-shot learning \cite{DTL},\textcolor{red}{HL: no need to cite this, which lower our contribution. you may want to say "similar to xxx, our method can be optimized xxx"} which aims to train a model and the model can be quickly adapted to a new task or new environment with a few shots of the new data. We can assume $\PHI$ is a pretrained model and $\PHI$ could reach the local minimal point with a few shots of data after a few steps of backward propagation. We highlight that we are not doing few-shot learning, but we are borrowing the idea and the objective of few-shot learning as the reasoning of doing the proposed approximation. 
In practice, $\hat{\PHI}$ can be obtained by several steps of backward propagation. If we consider one step only, the approximation of $\PHI^*$ is written as 
    \begin{equation} \label{eq-6}
        \PHI^* \approx \hat{\PHI} = \PHI -  \nabla_{\PHI} \mathcal{L}(x, y|\THETA, \PHI).
  	\end{equation}
As such, we relax the constraint of the original problem defined in Eq~\ref{eq-5}, and the gradient for updating $\THETA$ can be approximated as:
    \begin{equation} \label{eq-7}
        \begin{array}{cc}
      	    %& \nabla_{\THETA} \mathcal{L}(x, y|\THETA, \PHI^*) \approx \\ 
      	    %& \nabla_{\THETA} \mathcal{L}(x, y|\THETA, \PHI - \alpha %\nabla_{\PHI} \mathcal{L}(x, y|\THETA, \PHI)).
      	    \nabla_{\THETA} \mathcal{L}(x, y|\THETA, \PHI^*) \approx \nabla_{\THETA} \mathcal{L}(x, y|\THETA, \hat{\PHI}).
  		\end{array}
  	\end{equation}
By using this approximation, we avoid the complex high-order calculation of the original bi-level problem. 

% \textcolor{red}{HL: try to avoid using assume}
\begin{algorithm}[t]
        \SetAlgoLined
       
        1. The maximum training iterations $T$, approximation steps $K$ \\
        2. The learning rate $\alpha$ \\
        2. The training data of multiple data domains $D^S$ \\
        3. Initialize $\PHI$ and $\THETA$ 
       
        \For{t=1 to T}{
            Randomly split $D^S$ into $D_{\phi}$ and $D_\theta$ for the training $\PHI$ and $\THETA$ respectively. \\
            
            $\PHI^0 \leftarrow \PHI$ \\
            // Update $\PHI$ with $K$-shots of data\\
            \For{k=1 to K}{
                Sample mini-batch data $\{x^k_{\phi}, y^k_{\phi}\}$\\
                $\PHI^{k} = \PHI^{k-1} - \alpha \nabla_{\PHI} \mathcal{L}(x_{\phi}^k, y_{\phi}^k | \THETA, \PHI^{k-1})$
            }
            // Approximate the $\PHI^*$ by $\hat{\PHI}=\PHI^K$  \\
            $\hat{\PHI} \leftarrow \PHI^K$ \\
            Sample $\{x_{\theta}, y_{\theta}\}$ from $D_\theta$ \\
            Update $\THETA$ by the gradient of $ \nabla_{\THETA} \ \mathcal{L}(x_{\theta}, y_{\theta}|  \THETA, \hat{\PHI})$ \\
            Update $\PHI$ \ by \ $\PHI^K$\\
        }
    
    Output: the optimized $\PHI^*$ and $\THETA^*$
    
    \caption{Learning to extract Meta Pattern} \label{algo}
\end{algorithm}
        
In Algorithm \ref{algo}, we describe the overall optimization procedure for $\PHI$ and $\THETA$ based on Eq.\ref{eq-6} and Eq.\ref{eq-7}. As described in Algorithm \ref{algo}, there are two loops nested. In the outer loop,  source data from multiple domains $D^S$ is split into $D_{\phi}$ and $D_\theta$ and there is no domain overlapped between $D_{\phi}$ and $D_\theta$. As such, $D_{\phi}$ is ``unseen" to $\THETA$ to simulate $D^T$ as the data from unseen domains in each outer loop. Moreover, the overall optimization procedure is neat and completely end-to-end (single-stage). Complicated high-order gradient calculation is avoided. Moreover, the Meta Pattern Extractor $\PHI$ can be trained on the fly without using a surrogate model of $\THETA$ and the retraining of the target model $\THETA$ is not needed.  

For the instantiation of the Meta Pattern Extractor $\PHI$, we propose to parameterize $\PHI$ by a convolutional neural network because convolutional kernels can work as learnable filters used to extract features. The structure of $\PHI$ can be seen from Fig.~\ref{fig:phi}, which is a shallow network consisting of two convolutional layers. We do not consider a deep network because a deep network usually has more parameters and needs more data to fit. Meanwhile, in our approximation solution, the amount of data for the approximation is small. If $\PHI$ is deep, the approximation may be insufficient because of the small amount of data. Thus, we parameterize $\PHI$ by a convolutional network of two vanilla convolutional layers with $3\times3$ convolutional kernels. The sigmoid activation function is used at the end to constrain the output range from 0 to 1. Besides, the output MP has the same size as the input RGB image, which follows \cite{Pinto, FAS-MSR-TIFS-2019}. In the experiment section, we will discuss the effects of the other instantiations of $\PHI$ by considering central difference convolution \cite{CDCN-CVPR-2020}.

% The feature pyramid network
\subsection{Hierarchical Fusion Network}
The extracted MP provides task-aware discriminative information for FAS, while the original RGB image contains complete and detailed information. To effectively utilize both the detailed and discriminative information, we devise a two-stream Hierarchical Fusion Network (HFN) $\THETA$ to fuse the input RGB images and the extracted MP for FAS. As shown in \fig{fig:HFN}, the top stream of the HFN is the RGB stream that processes the information from RGB images, and the bottom stream is the MP stream that processes the information from the MP. The RGB and MP streams are identical. For implementation, we adopt the ResNet-50 \cite{ResNet}, which is commonly used in image classification, as the backbone for each stream. The features from the end of the two streams are fused to a fully connected layer to obtain a binary vector for the classification. Moreover, we improve the fusion performance by fusing the information more thoroughly from different feature hierarchies. As shown in \fig{fig:HFN}, the features from different feature hierarchies are fused progressively via our proposed Hierarchical Fusion Module (HFM). The HFM is inspired by the Feature Pyramid Network (FPN). While the FPN constructs the feature pyramid from different hierarchies progressively to improve the detection for small objects \cite{FPN}, we fuse information from different hierarchies progressively to improve the fusion.

At hierarchy $i$ of the HFN, an HFM $\mathcal{F}_i$ fuses feature maps from the top (RGB) stream $m^t_i$,  the bottom (MP) stream $m^{b}_{i}$, and the fusion result of the prior hierarchy $m_{i-1}$.  
The expression of $m_{i}$ can be written as 
    \begin{equation}\label{eq-8}
  		\begin{aligned}
	  	     m_{i} = & \mathcal{F}_i(m^t_i, m^b_i, m_{i-1})\\
	  	             = & \mathcal{C}(m^t_i) + \mathcal{C}(m^b_i) + \mathcal{U}[\mathcal{C}(m_{i-1})],
	  	    i=1,2,3, m_0 = \textbf{0},
  		\end{aligned}
  	\end{equation}
where $\mathcal{C}$ represents a convolutional layer that is used to align the number of channels of feature maps from different hierarchies, and $\mathcal{U}$ is the nearest interpolation function that upsamples $m_{i-1}$ to have the same size as $\mathcal{C}(m^b_i)$ and $\mathcal{C}(m^t_i)$ such that element-wise addition can be conducted. One similar fusion scheme that also fuses information from multiple feature hierarchies for the face anti-spoofing problem is the DC-CDN \cite{DCCDN-IJCAI-2021}. The difference between the fusion in our HFN and the fusion in the DC-CDN is that DC-CDN just directly concatenates the feature maps from multiple levels along the channel axis \cite{DCCDN-IJCAI-2021}, which is straightforward but could be coarse, while our HFN fuses information progressively. In the experiment section, we conduct the ablation study to compare our fusion of HFM and the fusion of concatenation.

In the final fusion hierarchy, the fused feature map is used to predict a $32\times32$ pixel map to take advantage of the pixel-wise supervision \cite{FAS-Auxiliary-CVPR-2018, deeppixel--ICB-2019, yu2021revisiting, Ternary-TIFS-2018}. As such, both the binary classification supervision and pixel-wise supervision are used to optimize our proposed HFN. We apply Binary Cross Entropy loss $\mathcal{L}_{BCE}$ to the binary vector output from the FC layer, where spoofing faces are labeled as ``0'' and genuine faces are labeled as ``1''. We apply pixel-wise Mean Square Error (MSE) loss $\mathcal{L}_{MSE}$ to the output pixel maps from the HFN. In the pixel-wise supervision, each spoofing face or genuine face is assigned a target $32\times32$ pixel map with all elements as ``0" or ``1", respectively. As such, the final loss for optimization is 

\begin{equation}\label{eq-9}
\mathcal{L} = \mathcal{L}_{BCE} +\mathcal{L}_{MSE}.
\end{equation}

\subsection{Testing}
In the testing stage, we combine the output binary vector and the pixel map to get the score and use the score for classification. As for an output binary vector $\textbf{s} = [s_0, s_1]$, where $s_0\in [0,1]$, $s_1\in[0,1]$, and $s_0+s_1=1$, and an output pixel map $m \in \{\mathbb{R^+}^{32\times32}|m_{(i,j)}\in [0, 1] \}$. The score used for the classification is calculated by
\begin{equation}\label{eq-10}
S = \frac{s_1 + mean(m)}{2}.
\end{equation}
The score $S \in [0, 1]$ represents the probability of the input being ``genuine".

% Meta-learning and overall algorithms

\section{Experiments} \label{Sec-4} \begin{figure}
    \centering
    \includegraphics[width=\linewidth]{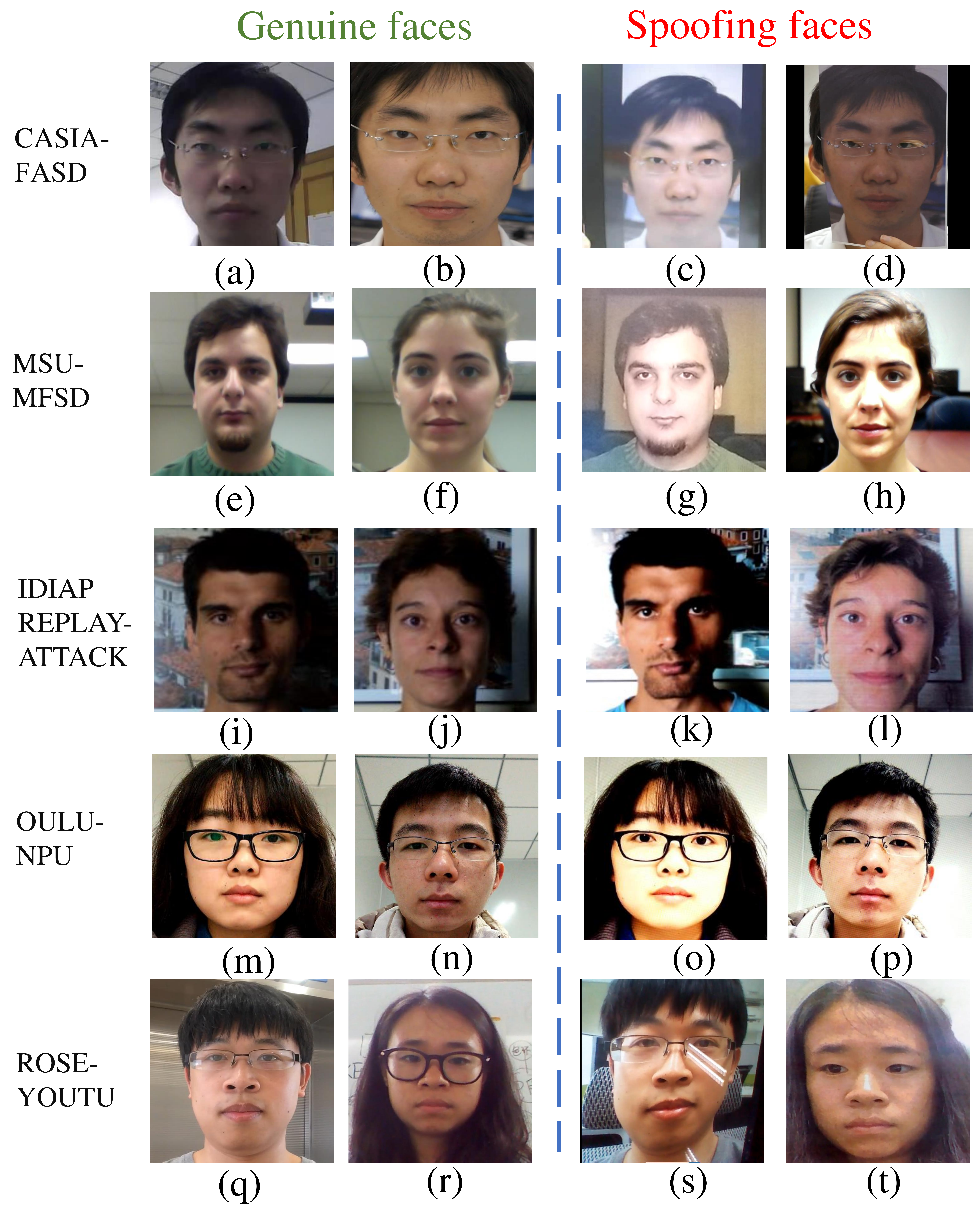}
    \caption{Genuine faces and spoofing faces examples from the CASIA-FASD dataset \cite{DB-CASIAFASD} (1-st row), the MSU-MFSD dataset \cite{LI-QUALITY} (2-nd row), the IDIAP REPLAY-ATTACK dataset \cite{MicroTexture-IJCB-2011} (3-rd row), the OULU-NPU dataset \cite{OULU_NPU_2017} (4-th row), and the ROSE-YOUTU \cite{FAS-UnsupervisedDA-TIFS-2018} (5-th row) dataset.}
    \label{fig:data}
\end{figure}
\begin{figure*}[ht]
    \centering
    \includegraphics[width=\linewidth]{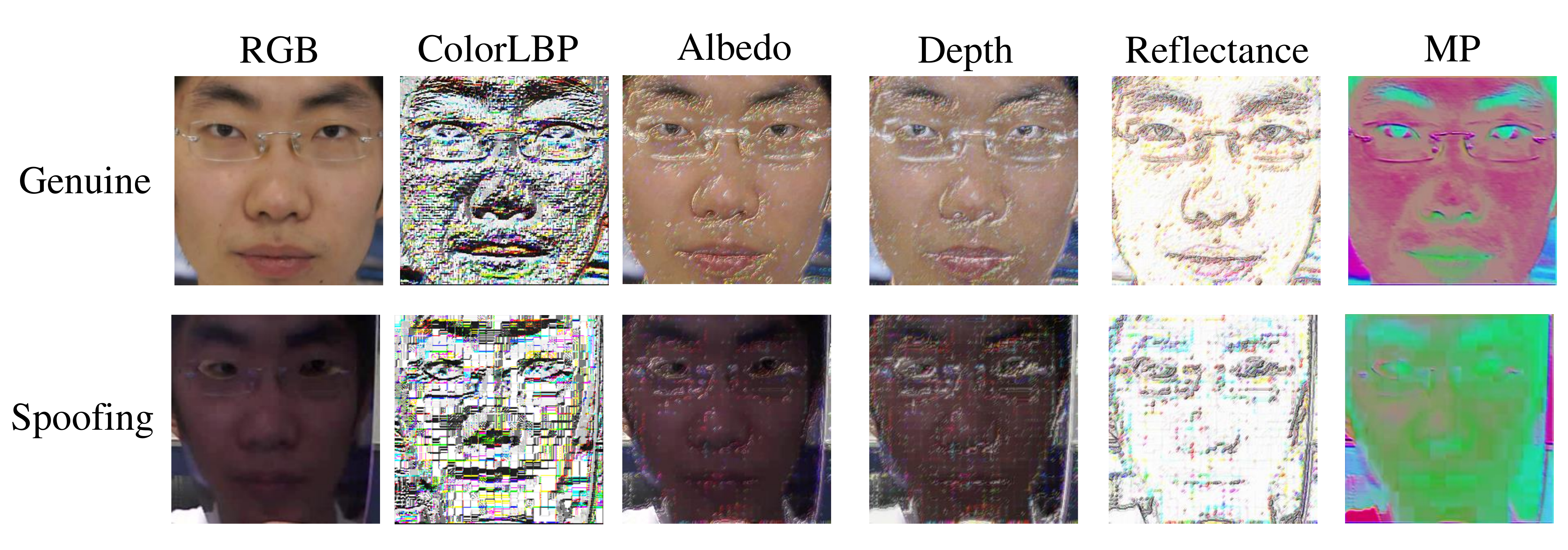}
    \caption{Visual comparison  between the ColorLBP map, Albedo map, Depth map, reflectance map, and our Meta Pattern (MP). The MP is extracted by the Meta Pattern Extracted trained in the ``O\&M\&I to C" experiment. The genuine and spoofing face examples are from the CASIA-MFSD dataset \cite{DB-CASIAFASD}. }
    \label{fig:vis}
\end{figure*}
\begin{table*}[tbp]
  \centering
  \caption{Comparison between different HFN with the MP and with the handcrafted feature maps on the MICO domain generalization benchmark. The performance metrics are HTER (\%) and AUC (\%). ``$\downarrow$" means the lower the better, and ``$\uparrow$" means the higher the better. }
    \begin{tabular}{|l|c|c|c|c|c|c|c|c|}
    \hline
    \multicolumn{1}{|c|}{\multirow{2}[4]{*}{Method}} & \multicolumn{2}{c|}{C\&I\&O to M} & \multicolumn{2}{c|}{O\&M\&I to C} & \multicolumn{2}{c|}{O\&C\&M to I} & \multicolumn{2}{c|}{I\&C\&M to O} \bigstrut\\
\cline{2-9}          & HTER $\downarrow$ & AUC $\uparrow$ & HTER $\downarrow$ & AUC $\uparrow$  & HTER $\downarrow$ & AUC $\uparrow$  & HTER $\downarrow$ & AUC $\uparrow$  \bigstrut\\
    \hline

    HFN+ColorLBP &   8.31    &   96.77    &     18.44  & 88.72      &  20.55     &   77.58    &   15.92    & 90.98  \bigstrut\\
    \hline
    
    HFN+Albedo & 8.70      &  96.46     &  20.89     &     87.69  &    21.55   &  76.44     & 16.78      &  88.58 \bigstrut\\
    \hline
    HFN+Depth &    7.36   &    96.90   &  18.67     &  89.60 & 18.50      &    80.40   &  15.95     & 89.99 \bigstrut\\
    \hline
    HFN+Reflectance &   5.65    &  \textbf{97.62}    &      18.22 & 88.57      & 20.90 & 77.23 &    17.16   & 88.98 \bigstrut\\
    \hline
    \hline
    HFN+MP (Ours) & \textbf{5.24} & 97.28 & \textbf{9.11} & \textbf{96.09} & \textbf{15.35} & \textbf{90.67} & \textbf{12.40} & \textbf{94.26} \bigstrut\\
    \hline
    \end{tabular}%
  \label{tab:handcraft}%
\end{table*}%

\subsection{Datasets}
For evaluation, we consider the situation of complicated source domain data and use the multi-source domain generalization setting introduced in \cite{MADDG-CVPR-2019}, which has been the FAS cross-domain benchmark used by the recent methods \cite{RFMetaFAS-AAAI-2020,NASFAS-TPAMI-2020, SSDG-CVPR-2020, MetaTeacher-TPAMI-2021}. In this setting, four benchmark datasets, the MSU-MFSD \cite{FAS-IDA-TIFS-2015} (\textbf{M}), IDIAP REPLAY-ATTACK (\textbf{I}) \cite{LBP-FAS-BIOSIG-2012}, CASIA-FASD \cite{DB-CASIAFASD} (\textbf{C}), and OULU-NPU \cite{OULU_NPU_2017} (\textbf{O}) datasets are used, where any three out of the four datasets are used for training and the left one is for testing. \textcolor{black}{For simplicity, we propose to refer this setting as \textbf{MICO} (with initial letters from the four datasets).}  We also consider the ROSE-YOUTU dataset \cite{FAS-UnsupervisedDA-TIFS-2018}, a dataset for industrial needs. Similar to the \textbf{MICO}, the ROSE-YOUTU (\textbf{Y}) is used to conduct other cross-domain experiments in \cite{DRUDA-TIFS-2020}, where the MSU-MFSD \cite{FAS-IDA-TIFS-2015}, IDIAP REPLAY-ATTACK (\textbf{I}) \cite{LBP-FAS-BIOSIG-2012}, and CASIA-FASD \cite{DB-CASIAFASD} (\textbf{C}) are also used in the experiment setting introduced in \cite{DRUDA-TIFS-2020}. For simplicity, we propose to refer the setting as \textbf{MICY}. We also use the \textbf{MICY} to provide a more extensive evaluation. Next, we will briefly introduce these 5 datasets.

\textbf{CASIA-FASD} \cite{DB-CASIAFASD} contains genuine and spoofing faces from 50 genuine subjects. Cameras of low, normal, and high imaging resolutions are used to capture face images. Therefore, each subject has 3 kinds of live faces captured under three different resolutions. Also, CASIA-FASD consists of three kinds of 2D attacks, warped photo-attack, cut photo attack, and video attacks. Thus, there are 3×3=9 kinds of spoofing faces. There are lighting variances but the variances are not annotated.

\textbf{IDIAP REPLAY-ATTACK} \cite{LBP-FAS-BIOSIG-2012} captures all genuine and spoofing faces from 50 genuine subjects. Five attack manners including four kinds of replayed faces and one kind of printed face are used to produce spoofing faces. The data is captured under two different lighting conditions: normal lighting and adverse lighting.

\textbf{MSU-MFSD} \cite{FAS-IDA-TIFS-2015} contains genuine and spoofing faces from 35 genuine subjects captured by two cameras. The production of spoofing faces is from two digital displays for replay attacks and from one printer for photo attacks. There is only indoor lighting condition to capturing data.

\textbf{OULU-NPU} \cite{OULU_NPU_2017} is a large-scale dataset with 55 subjects. The captured face images are of high image quality since the OULU-NPU dataset captures data of genuine faces and spoofing faces by six cameras of high resolution. The spoofing faces consist of two kinds of printed spoofing faces and two kinds of replayed spoofing faces. The data is collected under three environmental sessions.

\textbf{ROSE-YOUTU} (ROSE-YOUTU Face Liveness Detection Database) \cite{FAS-UnsupervisedDA-TIFS-2018} is a recent large-scale dataset from the industry. It collects data from 20 subjects. For each subject, there are 25 genuine and 150 to 200 spoofing face videos, captured by five kinds of camera modules, front-facing cameras of Hasee phone, Huawei phone, ZTE phone, iPad, and iPhone 5s, such that the resolutions range from $640 \times 480$ to $1280\times720$. Other than photo attacks and replay video attacks, the ROSE-YOUTU dataset involves various paper mask attacks. Moreover, ROSE-YOUTU diversely covers 5 different lighting conditions.
Some face examples  from these five datasets are shown in Fig.~\ref{fig:data}, and we can observe that different disparities can appear in different datasets. For example, \fig{fig:data}(d) shows the cutting edge. The color distortion appears in \fig{fig:data}(d), \fig{fig:data}(h), and \fig{fig:data}(o). \fig{fig:data}(o) shows moire patterns. \fig{fig:data}(s) shows reflection patterns.

\subsection{Implementations}
As for data processing, we use MTCNN \cite{MTCNN} to capture and crop face images from video frames. The captured face images are resized to $256\times256$ as the network input. To train networks, we use PyTorch \cite{pytorch} to implement the networks to conduct experiments.  We set the learning rate $\alpha=0.001$. We use the Stochastic Gradient Descent (SGD) optimizer with a momentum value of 0.9. Unless otherwise specified, we set the approximation steps $K=4$ (used in Algorithm \ref{algo}) for the approximation. In each mini-batch, we set the batch size as 4 for genuine/spoofing faces and for each dataset domain to balance the sample between each dataset domain and balance the ratio between genuine real and spoofing faces. As such, there will be a batch of 24 ($4\times2\times3$) if there are 3 source domain datasets. For performance evaluation and comparison, we follow \cite{RFMetaFAS-AAAI-2020,NASFAS-TPAMI-2020, SSDG-CVPR-2020, MetaTeacher-TPAMI-2021} to report Half Total Error Rate (HTER) and  Area Under the receptive operating characteristic Curve (AUC) as performance metrics.  
\begin{table*}[tbp]
  \centering
  \caption{Experimental results on the domain-generalization benchmarks (MICO). Our proposed method is compared with the state-of-the-art methods in terms of HTER (\%) and AUC (\%). ``$\downarrow$" means the lower the better, and ``$\uparrow$" means the higher the better.}
    \begin{tabular}{|l|c|c|c|c|c|c|c|c|}
    \hline
    \multirow{2}[4]{*}{Method} & \multicolumn{2}{c|}{C\&I\&O to M} & \multicolumn{2}{c|}{O\&M\&I to C} & \multicolumn{2}{c|}{O\&C\&M to I} & \multicolumn{2}{c|}{I\&C\&M to O} \\
\cline{2-9}          & HTER $\downarrow$ & AUC $\uparrow$ & HTER $\downarrow$ & AUC $\uparrow$  & HTER $\downarrow$ & AUC $\uparrow$ & HTER $\downarrow$ & AUC $\uparrow$  \\
\hline 
    MMD-AAE \cite{MMDAAE-CVPR-2018}   &  27.08 & 83.19 & 44.59 & 58.29 & 31.58 & 75.18 & 40.98 & 63.08\\
    \hline 
    MADDG \cite{MADDG-CVPR-2019}   & 17.69 & 88.06 & 24.5 & 84.51 & 22.19 & 84.99 & 27.98 & 80.02\\
    \hline
    RFMetaFAS \cite{RFMetaFAS-AAAI-2020}   & 13.89      &  93.98     &  20.27     &     88.16  &    17.30   &  90.48     &  16.45     &   91.16\\
    \hline
    NAS-Baesline w/ D-Meta \cite{NASFAS-TPAMI-2020} & 11.62 & 95.85 & 16.96 & 89.73 & 16.82 & 91.68 & 18.64 & 88.45 \\
    \hline
    NAS w/ D-Meta \cite{NASFAS-TPAMI-2020} & 16.85 & 90.42 & 15.21 & 92.64 & \textbf{11.63} & \textbf{96.98} & 13.16 & 94.18 \\
    \hline
    NAS-FAS \cite{NASFAS-TPAMI-2020} & 19.53 & 88.63 & 16.54 & 90.18 & 14.51 & 93.84 & 13.80 & 93.43 \\
    \hline
    SSDG-M \cite{SSDG-CVPR-2020}& 16.67 & 90.47 & 23.11 & 85.45 & 18.21 & 94.61 & 25.17 & 81.83 \\
    \hline
    SSDG-R  \cite{SSDG-CVPR-2020} & 7.38  & 97.17 & 10.44 & 95.94 & 11.71 & 96.59 & 15.61 & 91.54 \\
    \hline
    FAS-DR-BC(MT) \cite{MetaTeacher-TPAMI-2021}   &     11.67    &  93.09     &  18.44   &  89.67     &     11.93  &    94.95   &  16.23     & 91.18 \\
    \hline
    \hline
    HFN+MP (Ours)  & \textbf{5.24}  &	\textbf{97.28} &	\textbf{9.11} &	\textbf{96.09} &	15.35 &	90.67 & \textbf{12.40}&	\textbf{94.26} \\
    \hline
    \end{tabular}%
  \label{tab:sota-MICO}%
\end{table*}%
\begin{table*}[htbp]
  \centering
  \caption{Experimental results on the cross-domain benchmark MICY. The proposed method is compared with other methods in terms of HTER (\%) and AUC (\%). ``--'' means the results are not available. ``$\downarrow$" means the lower the better, and ``$\uparrow$" means the higher the better.}
    \begin{tabular}{|l|c|c|c|c|c|c|c|c|}
    \hline
    \multirow{2}[4]{*}{Method} & \multicolumn{2}{c|}{M\&C\&Y to I	
} & \multicolumn{2}{c|}{I\&C\&Y to M} & \multicolumn{2}{c|}{I\&M\&Y to C} & \multicolumn{2}{c|}{I\&C\&M to Y} \\
\cline{2-9}          & HTER $\downarrow$ & AUC $\uparrow$ & HTER $\downarrow$ & AUC $\uparrow$ & HTER $\downarrow$ & AUC $\uparrow$ & HTER $\downarrow$ & AUC $\uparrow$  \\
    \hline
    ADDA \cite{ADDA-CVPR-2017} \ \ \ \ \ \ \ \ \ \ \ \ \ \ \ \ \ \ \ \ \ \   &  35.40	& --&	34.20	& --&	33.40	&-- &	36.30 &
 \\
    \hline
    DRCN \cite{DRCN-ECCV} & 37.20	& --&	33.90 &-- &	31.40	& --&	35.70	 &--
 \\
    \hline
    DupGAN \cite{DupGan-CVPR-2018} & 38.10	&-- &	33.70	 &-- &	26.90	& --&	33.40 &--	\\
    \hline
    ADA \cite{ADA-ICB-2019} & 6.3	& --&	12.7 &-- &		37.8 &-- &		31.0 &-- \\
    \hline
    DR-UDA \cite{DRUDA-TIFS-2020} & 3.4	& --&	10.2 &-- &	20.4 &-- &		29.7 &--\\
    \hline
    \hline
    %HFN-R18 (Ours)  &    17.66 & 	81.12 &10.08 & 	95.93 & 15.94 & 93.08 & 21.43 & 	85.24 \\
    %HFN (Ours)  &     & 	 & & 	 &  &  & & 	 \\
    %HFN-R18+MP (Ours) & 17.66&	81.12 & 10.08&	95.93&	13.94&	93.08&	21.43&	85.24 \\
    %\hline
    HFN+MP (Ours) & 10.42&	95.58&	\textbf{7.31}&	96.79&	\textbf{9.44}&	96.05&	\textbf{17.24}&	89.76 \\
    \hline

    \end{tabular}%
  \label{tab-MICY-SOTA}%
\end{table*}%
\begin{table}[tbp]
\centering \small
\caption{Experimental results on the variant of domain-generalization benchmarks (MICO), where only two source datasets are used for training. The proposed method is compared with the state-of-the-art methods in terms of HTER (\%) and AUC (\%). ``$\downarrow$" means the lower the better, and ``$\uparrow$" means the higher the better.}
 
\begin{tabular}{|l|c|c|c|c|}
\hline
\multirow{2}{*}{Method} & \multicolumn{2}{c|}{M\&I to C} & \multicolumn{2}{c|}{M\&I to O}\\
\cline{2-5}
& HTER $\downarrow$ & AUC $\uparrow$& HTER $\downarrow$ & AUC $\downarrow$ \\
\hline
MS-LBP \cite{MicroTexture-IJCB-2011} & 51.16 & 52.09 & 43.63 & 58.07 \\
\hline
IDA \cite{FAS-IDA-TIFS-2015} & 45.16 & 58.80 & 54.52 & 42.17\\
\hline
ColorTexture \cite{FAS-ColorTexture-TIFS-2016} & 55.17 & 46.89 &  53.31 & 45.16 \\
\hline
LBP-TOP \cite{LBP-TOP-EJIVP-2014} & 45.27 &  54.88 & 47.26 & 50.21 \\
\hline
MADDG \cite{MADDG-CVPR-2019} & 41.02 & 64.33 & 39.35 & 65.10  \\
\hline
%DR-Net \cite{SSDG-CVPR-2020} & 31.67 & 75.23 & 34.02 & 72.65  \\
\hline 
SSDG-M \cite{SSDG-CVPR-2020} & 31.89 & 71.29 & 36.01 & 66.88  \\
\hline
\hline
 
%HFN-R18+MP(Ours) & 29.78&	73.37&	21.22&	85.97 \\
%\hline
HFN+MP (Ours) & \textbf{30.89} & \textbf{72.48} &	\textbf{20.94} & \textbf{86.71} \\
\hline
\end{tabular}%
\label{tab:MICO-2to1}%
\end{table}%
\subsection{Comparison with handcrafted features}
In this section, we compare the MP with the other handcrafted features. We follow \cite{FAS-ColorTexture-TIFS-2016} to extract the ColorLBP map from the R, G, and B channels respectively, using the parameters of the number of neighbor pixels $P=8$ and the Radius $R=1$. We retain the 2D spatial structure in the LBP map instead of collecting the histogram to construct LBP feature vectors. Besides, we extract the Albedo, Depth, and Reflectance maps by using the shape-of-shading algorithm \cite{Pinto} with the official code \footnote{https://github.com/allansp84/shape-from-shading-for-face-pad}. 

The extracted ColorLBP, Albedo, Depth, Reflectance, and MP maps are visually compared in \fig{fig:vis}. The first row and the second row are the genuine and spoofing face examples from the CASIA FASD dataset \cite{DB-CASIAFASD}. As for all these maps, we can observe the difference from these maps between the genuine and spoofing examples. As the other handcrafted features have been shown to provide discriminative information for the spoofing \cite{FAS-Survey-arXiv-2021}, it is difficult to compare and rank these maps to state which is more representative from the visual aspect. This difficulty corresponds to the motivating question of our work: ``how to choose desired handcrafted features to devise a more generalized hybrid method for the face anti-spoofing problem". Therefore, we address this question by using an end-to-end data-driven method: learning to extract the MP. 
To compare and analyze quantitatively, we train our proposed HFN with the MP based on Algorithm \ref{algo} (``HFN+MP"). To compare with the other handcrafted feature, we replace the MP with the ColorLBP, Albedo, Depth, and Reflectance maps respectively to train the other HFNs, denoted as ``HFN+ColorLBP", ``HFN+Albedo", ``HFN+Depth", and  ``HFN+Reflectance" respectively. 
The experimental results on the \textbf{MICO} setting  can be seen from \tab{tab:handcraft}. When we compare the handcrafted ColorLBP, Reflectance, Albedo, and Depth maps in \tab{tab:handcraft} (exclude MP), we can observe that each map has its own advantages and disadvantages in different experiments. For example, ``HFN+Reflectance" achieves the best HTER and AUC in ``C\&I\&O to M"; ``HFN+Depth" achieves the best HTER and AUC in ``O\&C\&M to I"; ``HFN+ColorLBP" achieves the best HTER and AUC in ``I\&C\&O to M". This observation shows that the representation ability of different handcrafted features are different in different source domain data, which reinforces our motivation that it is non-trivial to manually extract handcrafted features to improve models' generalization ability under different complicated source domain data. As such, we propose to learn a generalized model by learning to extract MP. We can see from \tab{tab:handcraft} that our ``HFN+MP'' can achieve the best AUC results in ``O\&M\&I to C", ``O\&C\&M to I", and ``I\&C\&M to O" and the best HTER results in all the four experiments. In summary, the experimental results in \tab{tab:handcraft} justify the motivation of our work and manifest the effectiveness of our proposed MP.
\subsection{Comparison with state-of-the-art methods}
\subsubsection{Comparisons on the MICO domain generalization benchmark}
Recent FAS methods about domain generalization are using the \textbf{MICO} benchmark for evaluation \cite{RFMetaFAS-AAAI-2020,NASFAS-TPAMI-2020, SSDG-CVPR-2020, MetaTeacher-TPAMI-2021}. We also follow the \textbf{MICO} benchmark to compare our method with state-of-the-art methods. The results are shown in \tab{tab:sota-MICO}. In the ``C\&I\&O to M", ``O\&M\&I to C", and ``C\&I\&M to O" settings, our method HFN+MP achieves the best HTER and AUC results. In short, our method can generally achieve state-of-the-art performance.
\begin{figure*}
    \centering
    \includegraphics[width=\linewidth]{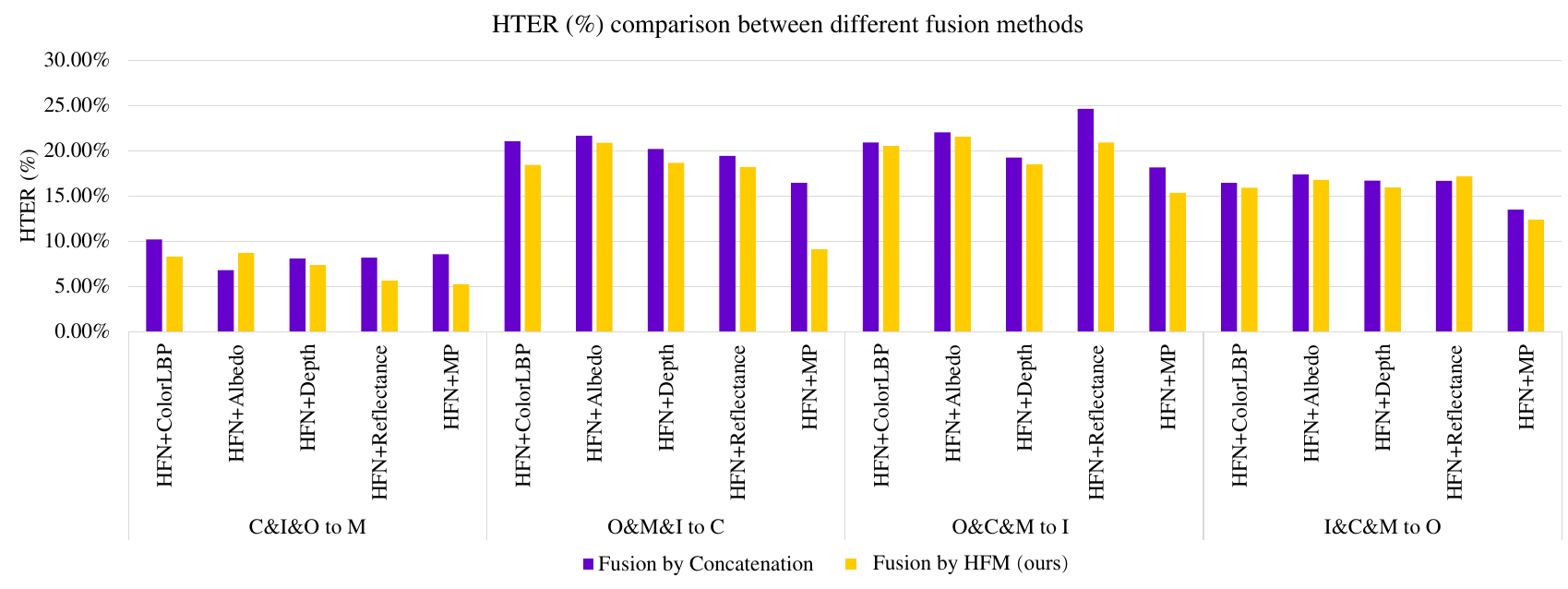}
    \caption{Performance comparisons between different methods of fusing features from different hierarchies in the HFN by using the MICO experiment setting (best viewed in color). The purple bars (left in each pair) show the HTER (\%) results of concatenating feature maps for fusion, and the yellow bars (right in pair) show the results of using our HFM for fusion. The lower the bar, the better the performance.}
    \label{fig:fusion_exp}
\end{figure*}
\begin{figure}
    \centering
    \includegraphics[width=\linewidth]{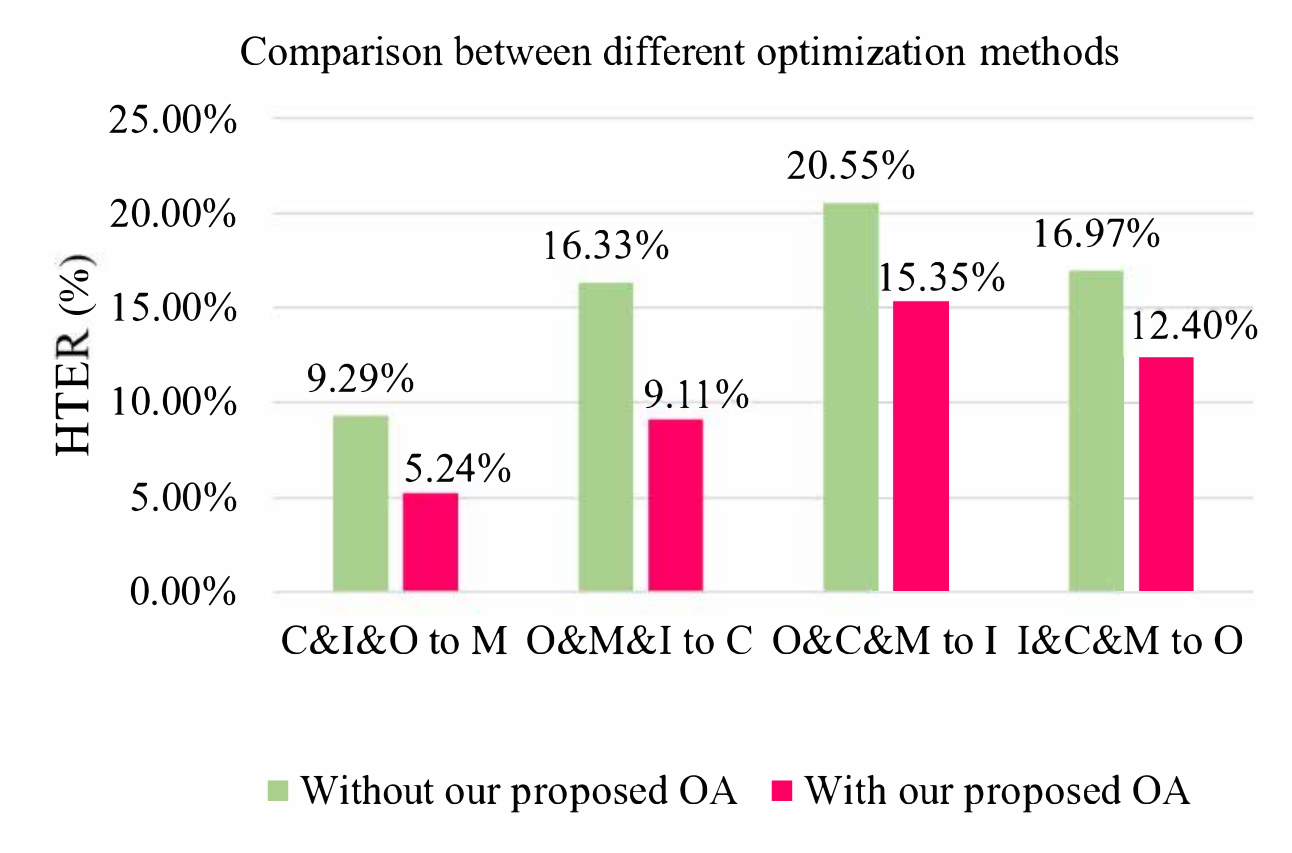}
    \caption{Performance comparisons between different optimization algorithms for $\PHI$ and $\THETA$ on the MICO benchmarks (best viewed in color). The green bars (left in each pair) show the HTER (\%) results of training $\PHI$ and $\THETA$ together by using ERM, without using our proposed Optimization Algorithm  (OA). The pink bars (right in each pair) show the results of training $\PHI$ and $\THETA$ with our proposed OA.}
    \label{fig:strategy}
\end{figure}
\begin{figure*}
    \centering
    \includegraphics[width=0.9\linewidth]{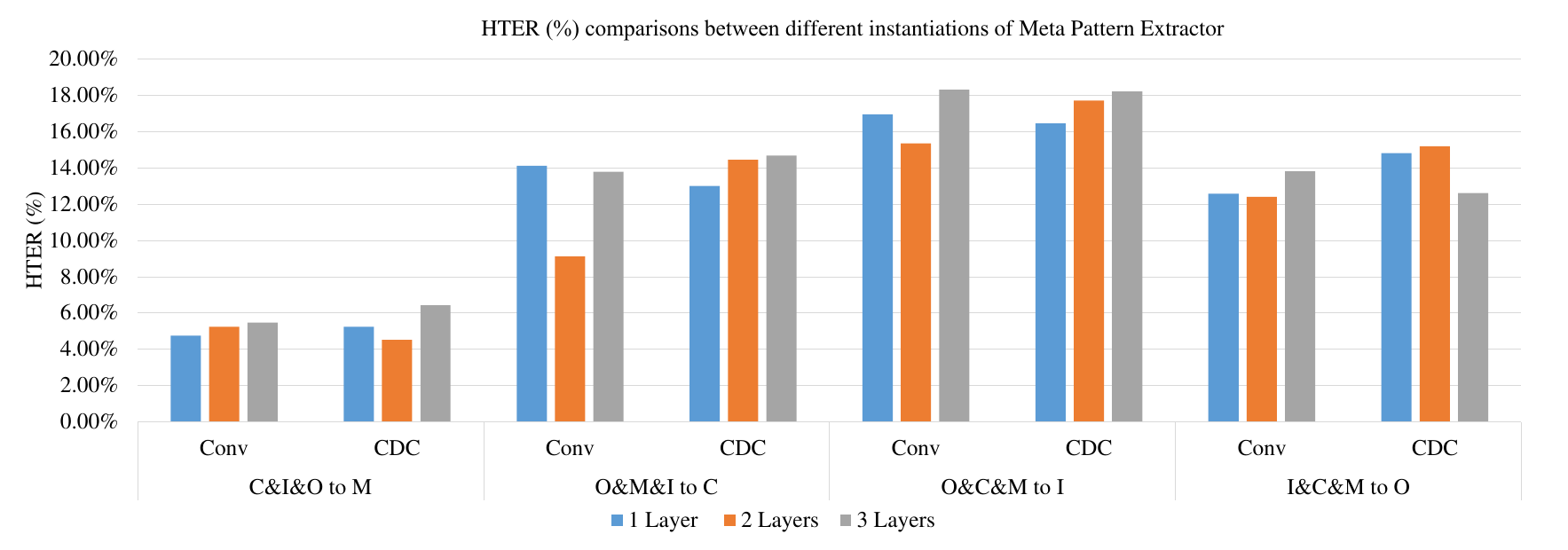}
    \caption{Performance comparisons between different instantiations of the Meta Pattern Extractors ($\PHI$) by using the MICO setting (best viewed in color). ``1 Layer Conv" means $\PHI^{CONV1}$, ``2 Layers Conv" means $\PHI^{CONV2}$, ``2 Layers CDC" means $\PHI^{CDC2}$,  \textit{etc.}}
    \label{fig:extractor}
\end{figure*}
\begin{figure}
    \centering
    \includegraphics[width=\linewidth]{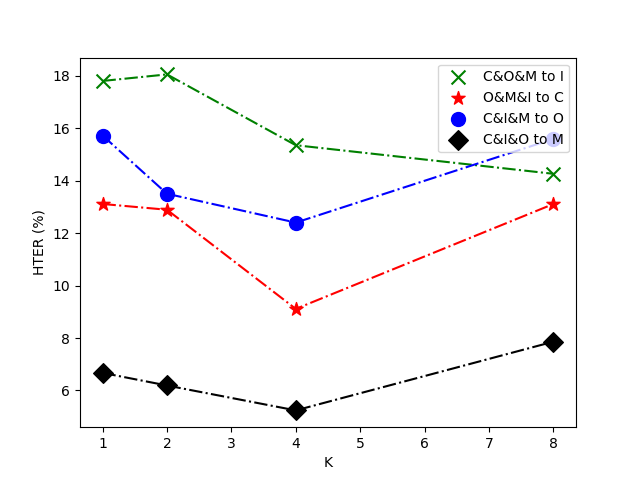}
    \caption{Performance comparisons between different steps  $K$ for the approximation. HTER(\%) results of $K=1,2,4,8$ are reported in the MICO benchmark.}
    \label{fig:k}
\end{figure}
\subsubsection{Comparisons in the MICY cross-domain experiments}
To further evaluate the proposed method, we follow \cite{DRUDA-TIFS-2020} to use the \textbf{MICY} setting to conduct domain generalization experiments. The experimental results can be found in \tab{tab-MICY-SOTA}. In \tab{tab-MICY-SOTA}, the listed methods are based on domain adaptation (DA), which uses the target domain data, but our method does not use target domain data. Even compared with these DA methods, our ``HFN+MP" can significantly achieve lower HTER than the other methods in ``I\&C\&Y to M", ``I\&M\&Y to C", and ``I\&C\&M to Y" settings. The results show the effectiveness of our MP. Although the AUC results of the other listed methods are not available, we still provide the AUC results of our method for readers' reference if they are interested to make comparisons.

\subsubsection{Limited source domains}
We also evaluate our proposed method when the number of source data domains is limited. We follow \cite{SSDG-CVPR-2020} to evaluate our proposed method with a variant of the MICO benchmark, where two datasets are used as source training data. As shown in \tab{tab:MICO-2to1}, our proposed method achieves the best HTER and AUC compared with state-of-the-art methods. Moreover, in the ``M\&I to O'' experiment, our proposed method shows over 10\% HTER lower than other methods. As such, our proposed method is still effective when fewer source data domains are available.

Overall, via extensive comparisons with state-of-the-art methods in multi-source domain generalization experiments, we manifest the effectiveness of our novel idea of training deep models by learning to extract Meta Pattern for the FAS problem. This provides a useful reference for future works about developing more generalized methods for the face anti-spoofing task, and maybe other tasks. 

\subsection{Ablation study}
In this section, we conduct the ablation study to present how each component of our proposed method can have influences on performance. We study the effectiveness of the fusion with HFM in the HFN. We also study the effectiveness of our approximation solution described in Algorithm \ref{algo} by comparing it to an end-to-end training strategy and trying different steps $K$ for approximation. Besides, we implement different instantiations of the Meta Pattern Extractor to observe the results. We conduct the ablation experiments by using the \textbf{MICO} benchmark and report the HTER results for comparison.
% Shows the effectiveness of HFN
% Our meta-learning results

\subsubsection{The effect of the proposed Hierarchical Fusion Module}
In our HFN, we adopt the Hierarchical Fusion Module to fuse the RGB images and the extracted MP from multiple feature hierarchies progressively. A two-stream network DC-CDN \cite{DCCDN-IJCAI-2021} for the FAS also fuses information from multiple feature hierarchies from the two streams. In the DC-CDN, the features maps from different hierarchies are directly concatenated along the channel axis, which is straightforward but could be coarse. We improve the fusion by using the HFM to progressively fuse the features. In this ablation experiment, we still use the two-stream structure of HFN, but we remove the HFM and concatenate the feature maps along the channel axis to do the fusion. The HTER results of different fusion methods are compared in \fig{fig:fusion_exp}. We can see that our HFM achieves lower HTER performance than the concatenation in 18 out of 20 pairs of experiments. Not only can our proposed HFM benefit the MP but also can be useful for other handcrafted feature maps.
\subsubsection{The effect of the proposed optimization algorithm}
According to Algorithm \ref{algo}, we optimize $\PHI$ and $\THETA$ separately. If we treat $\PHI$ and $\THETA$ as a whole that $\hat{\THETA}=[\THETA, \PHI ]$,  $\hat{\THETA}$ can be trained directly in an end-to-end manner according to Eq.~\ref{eq-1} and Eq.~\ref{eq-3}. In this ablation experiment, we study the effectiveness of our proposed Optimization Algorithm (OA). The ablation experimental results are shown in \fig{fig:strategy}. We can see that our OA can achieve lower HTER results than the direct end-to-end training of $\hat{\THETA}$ without our OA in the four experiments consistently. Therefore, the experimental results in \fig{fig:strategy} can show the effectiveness of our proposed OA.

\subsubsection{The effect of different steps $K$ for the approximation} 
In our proposed optimization algorithm (Algorithm~\ref{algo}), we approximate the local minima $\hat{\PHI}$ by doing $K$ steps of gradient descent. The quality of the approximation of $\hat{\PHI}$ is essential to the final optimization results. In this ablation experiment, we study how $K$ can affect the generalization performance. The HTER results of different $K$ (1, 2, 4, 8) are plotted in \fig{fig:k}. We can see from \fig{fig:k} that when $K$ increases from 1 to 4, the performance generally increases. We conjecture that when $K$ is small, the approximation is not sufficient enough. We observe that when $K$ increases from 4 to 8, the performance could drop. We conjecture the reason that when $K$ increases to a large number, the approximated local minima could get into an ``overfitting" pitfall and could lead to poorer performance.

\subsubsection{The effect of different Meta Pattern Extractors} 
In our proposed method, we parameterize $\PHI$ by a network of two convolutional layers, denoted as $\PHI^{CONV2}$. In this ablation experiment, we study the influence of different instantiations of $\PHI$ by adding or removing one layer to get a network $\PHI^{CONV1}$ of one convolutional layer and a network $\PHI^{CONV3}$ of three convolutional layers as the Meta Pattern Extractors to make comparisons with $\PHI^{CONV2}$. 
We also explore the central difference convolution (CDC) \cite{CDCN-CVPR-2020}, which is delicately designed for the face anti-spoofing problem. We replace the vanilla convolutions with the CDC in $\PHI^{CONV1}$, $\PHI^{CONV2}$, and $\PHI^{CONV3}$ respectively and obtain $\PHI^{CDC1}$, $\PHI^{CDC2}$, and $\PHI^{CDC3}$. As shown in \fig{fig:extractor}, different instantiations of $\PHI$ can have different performance. For example, in the experiment of ``C\&I\&O to M", $\PHI^{CONV2}$ has the same number of layers as $\PHI^{CDC2}$, but $\PHI^{CDC2}$ achieves lower HTER than $\PHI^{CONV2}$. However, $\PHI^{CDC1}$ and $\PHI^{CDC3}$ achieve higher HTER than the $\PHI^{CONV1}$ and $\PHI^{CONV3}$ respectively. We can empirically observe that different instantiations of $\PHI$ can influence the performance, but a theoretical analysis of the influence needs more effort to work out in the future. Also, the design of $\PHI$ is an open problem beyond the scope of this paper, and we will leave it for our future research.

\section{Conclusion and Future Work}\label{Sec-5}
In this paper, we present a novel method for the domain generalized FAS by learning to extract the Meta Pattern. Our method pushes the hybrid method one step further to be a fully end-to-end method by learning to extract the Meta Pattern, without extracting handcrafted features manually. Besides, we devise a novel two-stream Hierarchical Fusion Network with our proposed Hierarchical Fusion Module to fuse the information from the RGB images and the MP. Moreover, our method can achieve state-of-the-art performance in the MICO and MICY domain generalization benchmarks.

In the future, we will explore to develop more effective methods based on the idea of Meta Pattern, such as improving the Meta Pattern Extractor and the algorithm of optimization. We can also explore migrating the idea of Meta Pattern to other areas, such as deep fake detection.

% if have a single appendix:
%\appendix[Proof of the Zonklar Equations]
% or
%\appendix  % for no appendix heading
% do not use \section anymore after \appendix, only \section*
% is possibly needed

% use appendices with more than one appendix
% then use \section to start each appendix
% you must declare a \section before using any
% \subsection or using \label (\appendices by itself
% starts a section numbered zero.)
%
\bibliographystyle{ieeetr}
\bibliography{main.bib}

\begin{thebibliography}{10}

\bibitem{hashemifard2021compact}
S.~Hashemifard and M.~Akbari, ``A compact deep learning model for face spoofing
  detection,'' {\em arXiv preprint arXiv:2101.04756}, 2021.

\bibitem{FAS-ColorTexture-TIFS-2016}
Z.~{Boulkenafet}, J.~{Komulainen}, and A.~{Hadid}, ``{Face Spoofing Detection
  Using Colour Texture Analysis},'' {\em IEEE Transactions on Information
  Forensics and Security}, vol.~11, pp.~1818--1830, Aug 2016.

\bibitem{FAS-IDA-TIFS-2015}
D.~{Wen}, H.~{Han}, and A.~K. {Jain}, ``{Face Spoof Detection With Image
  Distortion Analysis},'' {\em IEEE Transactions on Information Forensics and
  Security}, vol.~10, pp.~746--761, April 2015.

\bibitem{moire-analysis-TIFS-2015}
D.~C. Garcia and R.~L. de~Queiroz, ``{Face-Spoofing 2D-Detection Based on
  Moiré-Pattern Analysis},'' {\em IEEE Transactions on Information Forensics
  and Security}, vol.~10, no.~4, pp.~778--786, 2015.

\bibitem{IQA-ICPR-2014}
J.~{Galbally} and S.~{Marcel}, ``{Face Anti-spoofing Based on General Image
  Quality Assessment},'' in {\em 2014 22nd International Conference on Pattern
  Recognition}, pp.~1173--1178, Aug 2014.

\bibitem{DoG-ECCV-2010}
X.~Tan, Y.~Li, J.~Liu, and L.~Jiang, ``{Face Liveness Detection from a Single
  Image with Sparse Low Rank Bilinear Discriminative Model},'' in {\em European
  Conference on Computer Vision}, pp.~504--517, 2010.

\bibitem{SURF-SPL-2017}
Z.~Boulkenafet, J.~Komulainen, and A.~Hadid, ``{Face Antispoofing Using
  Speeded-Up Robust Features and Fisher Vector Encoding},'' {\em IEEE Signal
  Processing Letters}, vol.~24, no.~2, pp.~141--145, 2017.

\bibitem{LBP-TOP-EJIVP-2014}
T.~de~Freitas~Pereira, J.~Komulainen, A.~Anjos, J.~M. De~Martino, A.~Hadid,
  M.~Pietik{\"a}inen, and S.~Marcel, ``Face liveness detection using dynamic
  texture,'' {\em EURASIP Journal on Image and Video Processing}, vol.~2014,
  no.~1, p.~2, 2014.

\bibitem{LBP-FAS-BIOSIG-2012}
I.~{Chingovska}, A.~{Anjos}, and S.~{Marcel}, ``On the effectiveness of local
  binary patterns in face anti-spoofing,'' in {\em 2012 BIOSIG - Proceedings of
  the International Conference of Biometrics Special Interest Group (BIOSIG)},
  pp.~1--7, Sep. 2012.

\bibitem{Motion-IJCB-2011}
A.~Anjos and S.~Marcel, ``Counter-measures to photo attacks in face
  recognition: {A} public database and a baseline,'' in {\em 2011 International
  Joint Conference on Biometrics (IJCB)}, pp.~1--7, 2011.

\bibitem{MotionLBP-ICB-2013}
J.~{Komulainen}, A.~{Hadid}, M.~{Pietikäinen}, A.~{Anjos}, and S.~{Marcel},
  ``Complementary countermeasures for detecting scenic face spoofing attacks,''
  in {\em 2013 International Conference on Biometrics (ICB)}, pp.~1--7, June
  2013.

\bibitem{FAS-3DCNN-TIFS-2018}
H.~{Li}, P.~{He}, S.~{Wang}, A.~{Rocha}, X.~{Jiang}, and A.~C. {Kot},
  ``{Learning Generalized Deep Feature Representation for Face
  Anti-Spoofing},'' {\em IEEE Transactions on Information Forensics and
  Security}, vol.~13, pp.~2639--2652, Oct 2018.

\bibitem{CAI-2020-DRL}
R.~Cai, H.~Li, S.~Wang, C.~Chen, and A.~C. Kot, ``{DRL-FAS: A Novel Framework
  Based on Deep Reinforcement Learning for Face Anti-Spoofing},'' {\em IEEE
  Transactions on Information Forensics and Security}, vol.~16, pp.~937--951,
  2020.

\bibitem{CDCN-CVPR-2020}
Z.~Yu, C.~Zhao, Z.~Wang, Y.~Qin, Z.~Su, X.~Li, F.~Zhou, and G.~Zhao,
  ``{Searching Central Difference Convolutional Networks for Face
  Anti-Spoofing},'' in {\em 2020 IEEE/CVF Conference on Computer Vision and
  Pattern Recognition (CVPR)}, pp.~5294--5304, 2020.

\bibitem{FAS-LSTMCNN-ICASSP-2018}
Z.~Sun, L.~Sun, and Q.~Li, ``Investigation in spatial-temporal domain for face
  spoof detection,'' in {\em 2018 IEEE International Conference on Acoustics,
  Speech and Signal Processing (ICASSP)}, pp.~1538--1542, 2018.

\bibitem{LIZHI}
Z.~Li, H.~Li, K.-Y. Lam, and A.~C. Kot, ``{Unseen Face Presentation Attack
  Detection with Hypersphere Loss},'' in {\em ICASSP 2020 - 2020 IEEE
  International Conference on Acoustics, Speech and Signal Processing
  (ICASSP)}, pp.~2852--2856, 2020.

\bibitem{Ternary-TIFS-2018}
W.~Sun, Y.~Song, C.~Chen, J.~Huang, and A.~C. Kot, ``{Face Spoofing Detection
  Based on Local Ternary Label Supervision in Fully Convolutional Networks},''
  {\em IEEE Transactions on Information Forensics and Security}, vol.~15,
  pp.~3181--3196, 2020.

\bibitem{CBL}
B.~Chen, W.~Yang, H.~Li, S.~Wang, and S.~Kwong, ``Camera invariant feature
  learning for generalized face anti-spoofing,'' {\em IEEE Transactions on
  Information Forensics and Security}, vol.~16, pp.~2477--2492, 2021.

\bibitem{FAS-UnsupervisedDA-TIFS-2018}
H.~{Li}, W.~{Li}, H.~{Cao}, S.~{Wang}, F.~{Huang}, and A.~C. {Kot},
  ``{Unsupervised Domain Adaptation for Face Anti-Spoofing},'' {\em IEEE
  Transactions on Information Forensics and Security}, vol.~13, pp.~1794--1809,
  July 2018.

\bibitem{FAS-Survey-arXiv-2021}
Z.~Yu, Y.~Qin, X.~Li, C.~Zhao, Z.~Lei, and G.~Zhao, ``{Deep Learning for Face
  Anti-Spoofing: A Survey},'' {\em arXiv preprint arXiv:2106.14948}, 2021.

\bibitem{FAS-Auxiliary-CVPR-2018}
Y.~Liu, A.~Jourabloo, and X.~Liu, ``{Learning Deep Models for Face
  Anti-Spoofing: Binary or Auxiliary Supervision},'' in {\em Proceedings of the
  IEEE Conference on Computer Vision and Pattern Recognition}, (Salt Lake City,
  UT), pp.~389--398, 2018.

\bibitem{DTL}
Y.~Liu, J.~Stehouwer, A.~Jourabloo, and X.~Liu, ``{Deep Tree Learning for
  Zero-Shot Face Anti-Spoofing},'' in {\em 2019 IEEE/CVF Conference on Computer
  Vision and Pattern Recognition (CVPR)}, pp.~4675--4684, 2019.

\bibitem{MicroTexture-IJCB-2011}
J.~{Määttä}, A.~Hadid, and M.~{Pietikäinen}, ``Face spoofing detection from
  single images using micro-texture analysis,'' in {\em 2011 International
  Joint Conference on Biometrics (IJCB)}, pp.~1--7, 2011.

\bibitem{HOG}
J.~Komulainen, A.~Hadid, and M.~Pietikäinen, ``Context based face
  anti-spoofing,'' in {\em 2013 IEEE Sixth International Conference on
  Biometrics: Theory, Applications and Systems (BTAS)}, pp.~1--8, 2013.

\bibitem{deep-forest-lbp}
R.~Cai and C.~Chen, ``Learning deep forest with multi-scale local binary
  pattern features for face anti-spoofing,'' {\em arXiv preprint
  arXiv:1910.03850}, 2019.

\bibitem{moire-icb}
K.~Patel, H.~Han, A.~K. Jain, and G.~Ott, ``Live face video vs. spoof face
  video: Use of moiré patterns to detect replay video attacks,'' in {\em 2015
  International Conference on Biometrics (ICB)}, pp.~98--105, 2015.

\bibitem{LI-QUALITY}
H.~{Li}, S.~{Wang}, and A.~C. {Kot}, ``Face spoofing detection with image
  quality regression,'' in {\em 2016 Sixth International Conference on Image
  Processing Theory, Tools and Applications (IPTA)}, pp.~1--6, 2016.

\bibitem{FAS-CNN-ComputerScience-2014}
J.~Yang, Z.~Lei, and S.~Z. Li, ``{Learn Convolutional Neural Network for Face
  Anti-Spoofing},'' {\em Computer Science}, vol.~9218, pp.~373--384, 2014.

\bibitem{vggnet}
K.~Simonyan and A.~Zisserman, ``Very deep convolutional networks for
  large-scale image recognition,'' in {\em 3rd International Conference on
  Learning Representations, {ICLR} 2015, San Diego, CA, USA, May 7-9, 2015,
  Conference Track Proceedings} (Y.~Bengio and Y.~LeCun, eds.), 2015.

\bibitem{deeppixel--ICB-2019}
A.~George and S.~Marcel, ``{Deep Pixel-wise Binary Supervision for Face
  Presentation Attack Detection},'' in {\em 2019 International Conference on
  Biometrics (ICB)}, pp.~1--8, 2019.

\bibitem{yu2021revisiting}
Z.~Yu, X.~Li, J.~Shi, Z.~Xia, and G.~Zhao, ``{Revisiting Pixel-Wise Supervision
  for Face Anti-Spoofing},'' {\em IEEE Transactions on Biometrics, Behavior,
  and Identity Science}, vol.~3, no.~3, pp.~285--295, 2021.

\bibitem{NASFAS-TPAMI-2020}
Z.~Yu, J.~Wan, Y.~Qin, X.~Li, S.~Z. Li, and G.~Zhao, ``{NAS-FAS: Static-Dynamic
  Central Difference Network Search for Face Anti-Spoofing},'' {\em IEEE
  Transactions on Pattern Analysis and Machine Intelligence}, vol.~43, no.~9,
  pp.~3005--3023, 2021.

\bibitem{FAS-MSR-TIFS-2019}
H.~Chen, G.~Hu, Z.~Lei, Y.~Chen, N.~M. Robertson, and S.~Z. Li,
  ``{Attention-Based Two-Stream Convolutional Networks for Face Spoofing
  Detection},'' {\em IEEE Transactions on Information Forensics and Security},
  vol.~15, pp.~578--593, 2020.

\bibitem{TIFS-2019-MotionBlur}
L.~Li, Z.~Xia, A.~Hadid, X.~Jiang, H.~Zhang, and X.~Feng, ``{Replayed Video
  Attack Detection Based on Motion Blur Analysis},'' {\em IEEE Transactions on
  Information Forensics and Security}, vol.~14, no.~9, pp.~2246--2261, 2019.

\bibitem{Pinto}
A.~Pinto, S.~Goldenstein, A.~Ferreira, T.~Carvalho, H.~Pedrini, and A.~Rocha,
  ``{Leveraging Shape, Reflectance and Albedo From Shading for Face
  Presentation Attack Detection},'' {\em IEEE Transactions on Information
  Forensics and Security}, vol.~15, pp.~3347--3358, 2020.

\bibitem{CNN-LBPTOP-2017}
M.~Asim, Z.~Ming, and M.~Y. Javed, ``{CNN based spatio-temporal feature
  extraction for face anti-spoofing},'' in {\em 2017 2nd International
  Conference on Image, Vision and Computing (ICIVC)}, pp.~234--238, 2017.

\bibitem{rehman2019perturbing}
Y.~A.~U. Rehman, L.-M. Po, M.~Liu, Z.~Zou, and W.~Ou, ``{Perturbing
  Convolutional Feature Maps with Histogram of Oriented Gradients for Face
  Liveness Detection},'' in {\em International Joint Conference: 12th
  International Conference on Computational Intelligence in Security for
  Information Systems (CISIS 2019) and 10th International Conference on
  EUropean Transnational Education (ICEUTE 2019)}, pp.~3--13, Springer, 2019.

\bibitem{LI-DISTILLATION}
H.~{Li}, S.~{Wang}, P.~{He}, and A.~{Rocha}, ``Face anti-spoofing with deep
  neural network distillation,'' {\em IEEE Journal of Selected Topics in Signal
  Processing}, vol.~14, no.~5, pp.~933--946, 2020.

\bibitem{RFMetaFAS-AAAI-2020}
R.~Shao, X.~Lan, and P.~C. Yuen, ``{Regularized Fine-Grained Meta Face
  Anti-Spoofing},'' {\em Proceedings of the AAAI Conference on Artificial
  Intelligence}, vol.~34, pp.~11974--11981, Apr. 2020.

\bibitem{MetaTeacher-TPAMI-2021}
Y.~Qin, Z.~Yu, L.~Yan, Z.~Wang, C.~Zhao, and Z.~Lei, ``{Meta-teacher for Face
  Anti-Spoofing},'' {\em IEEE Transactions on Pattern Analysis and Machine
  Intelligence}, vol.~Early Access, pp.~1--1, 2021.

\bibitem{MADDG-CVPR-2019}
R.~Shao, X.~Lan, J.~Li, and P.~C. Yuen, ``{Multi-Adversarial Discriminative
  Deep Domain Generalization for Face Presentation Attack Detection},'' in {\em
  2019 IEEE/CVF Conference on Computer Vision and Pattern Recognition (CVPR)},
  pp.~10015--10023, 2019.

\bibitem{ResNet}
K.~He, X.~Zhang, S.~Ren, and J.~Sun, ``Deep residual learning for image
  recognition,'' in {\em Proceedings of the IEEE Conference on Computer Vision
  and Pattern Recognition}, pp.~770--778, 2016.

\bibitem{FPN}
T.-Y. Lin, P.~Dollár, R.~Girshick, K.~He, B.~Hariharan, and S.~Belongie,
  ``{Feature Pyramid Networks for Object Detection},'' in {\em 2017 IEEE
  Conference on Computer Vision and Pattern Recognition (CVPR)}, pp.~936--944,
  2017.

\bibitem{DCCDN-IJCAI-2021}
Z.~Yu, Y.~Qin, H.~Zhao, X.~Li, and G.~Zhao, ``{Dual-Cross Central Difference
  Network for Face Anti-Spoofing},'' in {\em 2021 International Joint
  Conference on Artificial Intelligence (IJCAI)}, 2021.

\bibitem{DB-CASIAFASD}
Z.~Zhang, J.~Yan, S.~Liu, Z.~Lei, D.~Yi, and S.~Z. Li, ``A face anti-spoofing
  database with diverse attacks,'' in {\em IAPR International Conference on
  Biometrics}, pp.~26--31, 2012.

\bibitem{OULU_NPU_2017}
Z.~Boulkenafet, J.~Komulainen, L.~Li, X.~Feng, and A.~Hadid, ``{OULU-NPU}: A
  mobile face presentation attack database with real-world variations,'' in
  {\em IEEE International Conference on Automatic Face and Gesture
  Recognition}, May 2017.

\bibitem{SSDG-CVPR-2020}
Y.~Jia, J.~Zhang, S.~Shan, and X.~Chen, ``{Single-Side Domain Generalization
  for Face Anti-Spoofing},'' in {\em 2020 IEEE/CVF Conference on Computer
  Vision and Pattern Recognition (CVPR)}, pp.~8481--8490, 2020.

\bibitem{DRUDA-TIFS-2020}
G.~Wang, H.~Han, S.~Shan, and X.~Chen, ``{Unsupervised Adversarial Domain
  Adaptation for Cross-Domain Face Presentation Attack Detection},'' {\em IEEE
  Transactions on Information Forensics and Security}, vol.~16, pp.~56--69,
  2021.

\bibitem{MTCNN}
K.~Zhang, Z.~Zhang, Z.~Li, and Y.~Qiao, ``{Joint Face Detection and Alignment
  Using Multitask Cascaded Convolutional Networks},'' {\em IEEE Signal
  Processing Letters}, vol.~23, no.~10, pp.~1499--1503, 2016.

\bibitem{pytorch}
A.~Paszke, S.~Gross, F.~Massa, A.~Lerer, J.~Bradbury, G.~Chanan, T.~Killeen,
  Z.~Lin, N.~Gimelshein, L.~Antiga, A.~Desmaison, A.~Kopf, E.~Yang, Z.~DeVito,
  M.~Raison, A.~Tejani, S.~Chilamkurthy, B.~Steiner, L.~Fang, J.~Bai, and
  S.~Chintala, ``Pytorch: An imperative style, high-performance deep learning
  library,'' in {\em Advances in Neural Information Processing Systems 32}
  (H.~Wallach, H.~Larochelle, A.~Beygelzimer, F.~d\textquotesingle
  Alch\'{e}-Buc, E.~Fox, and R.~Garnett, eds.), pp.~8024--8035, Curran
  Associates, Inc., 2019.

\bibitem{MMDAAE-CVPR-2018}
H.~Li, S.~J. Pan, S.~Wang, and A.~C. Kot, ``{Domain Generalization with
  Adversarial Feature Learning},'' in {\em Proceedings of the IEEE Conference
  on Computer Vision and Pattern Recognition}, pp.~5400--5409, 2018.

\bibitem{ADDA-CVPR-2017}
E.~Tzeng, J.~Hoffman, K.~Saenko, and T.~Darrell, ``Adversarial discriminative
  domain adaptation,'' in {\em Proceedings of the IEEE Conference on Computer
  Vision and Pattern Recognition}, pp.~7167--7176, 2017.

\bibitem{DRCN-ECCV}
M.~Ghifary, W.~B. Kleijn, M.~Zhang, D.~Balduzzi, and W.~Li, ``Deep
  reconstruction-classification networks for unsupervised domain adaptation,''
  in {\em European Conference on Computer Vision}, pp.~597--613, Springer,
  2016.

\bibitem{DupGan-CVPR-2018}
L.~Hu, M.~Kan, S.~Shan, and X.~Chen, ``Duplex generative adversarial network
  for unsupervised domain adaptation,'' in {\em Proceedings of the IEEE
  Conference on Computer Vision and Pattern Recognition}, pp.~1498--1507, 2018.

\bibitem{ADA-ICB-2019}
G.~Wang, H.~Han, S.~Shan, and X.~Chen, ``Improving cross-database face
  presentation attack detection via adversarial domain adaptation,'' in {\em
  2019 International Conference on Biometrics (ICB)}, pp.~1--8, IEEE, 2019.

\end{thebibliography}

%\appendices
%\section{Proof of the First Zonklar Equation}

% you can choose not to have a title for an appendix
% if you want by leaving the argument blank
%\section{}
%Appendix two text goes here.

% use section* for acknowledgment

% Can use something like this to put references on a page
% by themselves when using endfloat and the captionsoff option.
\ifCLASSOPTIONcaptionsoff
  \newpage
\fi

% that's all folks
\end{document}